\documentclass[11pt]{article}

\usepackage[preprint]{acl}

\usepackage{times}
\usepackage{latexsym}

\usepackage[T1]{fontenc}
\usepackage[utf8]{inputenc}

\usepackage{microtype}

\usepackage{inconsolata}

\usepackage{graphicx}
\usepackage{amsmath}
\usepackage{booktabs}
\usepackage{longtable}
\usepackage{multirow}
\usepackage{CJKutf8}
\usepackage{algorithm}
\usepackage{algpseudocode}
\usepackage{dsfont}
\usepackage{tcolorbox}
\tcbuselibrary{skins} %preamble
\tcbuselibrary{breakable}
\usepackage{colortbl}
\definecolor{think}{RGB}{84, 130, 53}
\definecolor{search}{RGB}{191, 134, 179}
\definecolor{info}{RGB}{50, 100, 151}
\definecolor{translation}{RGB}{122, 183, 179}
\definecolor{text}{RGB}{69, 53, 193}
\usepackage{amssymb}
\usepackage{cuted} 
\usepackage{caption}
\usepackage{subcaption}
\usepackage{tikz}
\usepackage{pifont}
\usepackage{array}
\usepackage{makecell}

\definecolor{myred}{RGB}{255,0,0}

\newtheorem{definition}{Definition}

\newtcbox{\clustertab}[1]{on line, box align=base, colback={#1},colframe={#1},size=fbox,arc=2pt,top=-1.5pt, bottom=-1.5pt, left=-1.5pt, right=-1.5pt, boxrule=0pt, enlarge left by=1pt}

\definecolor{CustomBlue}{RGB}{57,83,191}

\usepackage{arydshln}
\makeatletter
\def\adl@drawiv#1#2#3{%
        \hskip.5\tabcolsep
        \xleaders#3{#2.5\@tempdimb #1{1}#2.5\@tempdimb}%
                #2\z@ plus1fil minus1fil\relax
        \hskip.5\tabcolsep}
\newcommand{\cdashlinelr}[1]{%
  \noalign{\vskip 1.3pt
           \global\let\@dashdrawstore\adl@draw
           \global\let\adl@draw\adl@drawiv}
  \cdashline{#1}[.4pt/2pt]
  \noalign{\global\let\adl@draw\@dashdrawstore
           \vskip 3pt}}
\makeatother

\newcommand{\thinkblock}[1]{\textcolor{think}{\texttt{#1}}}
\newcommand{\searchblock}[1]{\textcolor{search}{\texttt{#1}}}
\newcommand{\infoblock}[1]{\textcolor{info}{\texttt{#1}}}
\newcommand{\translationblock}[1]{\textcolor{translation}{\texttt{#1}}}
\newcommand{\textblock}[1]{\textcolor{text}{\texttt{ #1 }}}

\newcommand{\se}{\mathcal{E}}

\usetikzlibrary{bending} 
\usetikzlibrary{arrows.meta}
\DeclareRobustCommand{\mycircletikz}{%
    \tikz[baseline=-1ex, line width=1.5pt, scale=0.15, rotate=55]{
        \draw[-{Triangle[length=1.1mm, width=1.3mm, bend]}, line cap=round] 
            (0,0.8) arc[start angle=90, delta angle=-160, radius=0.8];
        \draw[-{Triangle[length=1.1mm, width=1.3mm, bend]}, line cap=round] 
            (0,-0.8) arc[start angle=270, delta angle=-160, radius=0.8];
    }%
}

\newcommand{\subtxt}[1]{\text{#1}}

\newcommand{\gptfivename}{\texttt{gpt-5-2025-08-07}}
\newcommand{\datasetname}{Neko} % new dataset name
\newcommand{\modelname}{NeoAMT}

\title{\modelname{}: Neologism-Aware Agentic Machine Translation with Reinforcement Learning}

\author{Zhongtao Miao$^{1}$,
        Kaiyan Zhao$^{1}$,
        Masaaki Nagata$^{2}$,
        Yoshimasa Tsuruoka$^{1}$\\
$^{1}$The University of Tokyo \\
$^{2}$NTT Communication Science Laboratories, NTT, Inc. \\
\texttt{\{miao, kaiyan1006, tsuruoka\}@logos.t.u-tokyo.ac.jp} \\
\texttt{masaaki.nagata@ntt.com}
}

\begin{document}
\begin{CJK*}{UTF8}{gkai}
\maketitle
\begin{abstract}
Neologism-aware machine translation\footnote{``Neologism-aware machine translation'' is used here to denote translating source sentences containing neologisms, not translating neologisms individually at the word level. 
Regarding the use of the term ``neologism'', we follow the Wiktionary definition:~\url{https://en.wiktionary.org/wiki/Wiktionary:Neologisms}.} aims to translate source sentences containing neologisms into target languages.
This field remains underexplored compared with general machine translation (MT).
In this paper, we propose an agentic framework, \modelname{}, for neologism-aware machine translation equipped with a Wiktionary-based search toolkit.
Specifically, we first construct a dedicated dataset for neologism-aware machine translation and build a search toolkit grounded in Wiktionary.
The dataset covers 16 languages and 75 translation directions in total, derived from approximately 10 million records of an English Wiktionary dump.
The retrieval corpus of the search toolkit is also constructed from around 3 million cleaned records of the same dump.
We then leverage the dataset and toolkit to train a translation agent via reinforcement learning (RL) and to evaluate the accuracy of neologism-aware machine translation.
Furthermore, we propose an RL training framework featuring a novel reward design and an adaptive rollout generation strategy that exploits ``translation difficulty'' to further improve the translation quality of translation agents using our search toolkit\footnote{The code, dataset and search toolkit will be available at~\url{https://github.com/gpgg/neoamt}.}.
\end{abstract}

\section{Introduction}
\begin{figure}[ht]
    \centering
    \includegraphics[width=0.42\textwidth]{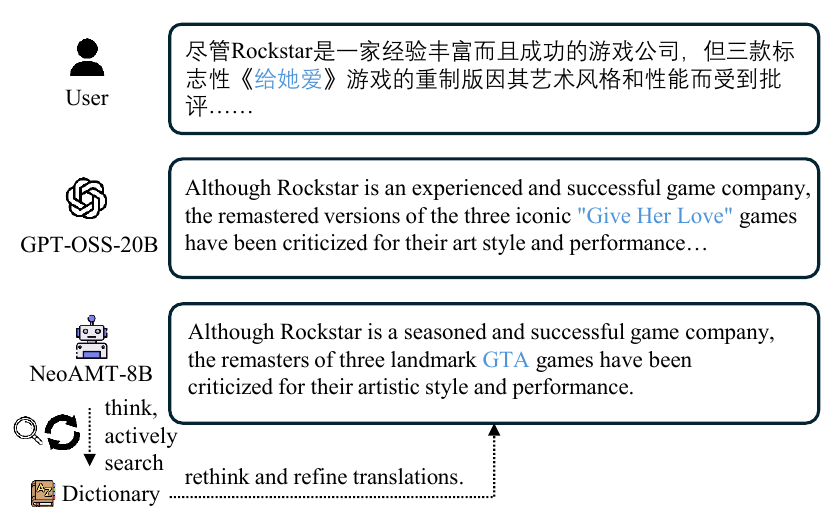}
    \caption{A Chinese-English neologism-aware machine translation example from our \datasetname{} dataset. The text in blue highlights a neologism or its translation. 
    The source text in Chinese contains a neologism ``给她爱'' which in this context refers to ``GTA'', an action-adventure video game series, rather than carrying its literal surface meaning of ``giving her love''. 
    % Our agentic model imitates the human translation process to actively search and refine translations in a multi-turn way when encountering neologisms or unknown words. 
    The circle \mycircletikz{} denotes that the thinking and searching processes can be repeated.
    The detailed thinking and searching process of our model can be found in Table~\ref{tab:find_new_meaning_example_our_model_gta}, Appendix~\ref{app:ex}.}
    \label{fig:translation_example}
\end{figure}
% With the development of large language models (LLMs), the machine translation models have shifted from the traditional Transfomer encoder-decoder architecture to large language models and the training paradigm also changed. 
% \kaiyan{introduction的部分可以更加突出我们的方法的优势}

Recently, reasoning-based language models~\citep{2409_openai_o1,Guo2025-qv,yang2025qwen3,team2025kimi} have gained a lot of attention because of the power of test-time scaling~\citep{snell2025scaling,muennighoff-etal-2025-s1,li-etal-2026-test}. Various attempts have been made to explore how to use large language models (LLMs) and reasoning with reinforcement learning (RL) on machine translation (MT)~\citep{feng-etal-2025-mt-r1,li2025tat,yang2025ssr,wang-etal-2025-drt,wang2025deep}. These studies mainly focus on the reward design to improve the MT performance of LLMs by leveraging their reasoning ability elicited by prompts and parametric knowledge. For example, MT-R1-Zero~\citep{feng-etal-2025-mt-r1} introduces a mixed reward mechanism. DeepTrans~\citep{wang2025deep} leverages DeepSeek-V3~\citep{liu2024deepseek} to design tailored format, thought, and translation rewards to enhance overall translation quality.

% SSR-Zero~\citep{yang2025ssr} uses self-evaluations as rewards. 
% TAT-R1~\citep{li2025tat} incorporates word alignment rewards into the final rewards. 

However, language is constantly evolving~\citep{doi:10.1073/pnas.96.14.8028}, driven by various factors such as cultural exchange, technological advancement, migration and generational change. People coin new words while the internal parametric knowledge in LLM-based neural machine translation systems remains frozen.
% these words usually play a crucial role in conveying the main meaning of the original sentences.
Thus, they cannot translate relatively new words well only using the reasoning ability and parametric knowledge, as shown in Figure~\ref{fig:translation_example}.

To tackle the aforementioned problem, we investigate how to enhance LLM performance on neologism-aware machine translation through a range of approaches, including supervised fine-tuning (SFT), RL with reasoning, and teaching LLMs how to consult dictionaries in an agentic way.
% agentic dictionary lookup.
Our contributions can be summarized as follows:
\begin{itemize}
    \item To the best of our knowledge, our research represents the first effort in exploring the effectiveness of neologism-aware machine translation agents by leveraging the interleaved reasoning and searching process with a dictionary.
    \item We propose a novel RL training framework, \modelname{}, based on GRPO~\citep{shao2024deepseekmath} which consists of a novel reward design and an adaptive sampling approach using the proposed translation difficulty metric.
    % \item For training and evaluation, we build a new dataset for neologism-aware machine translation and develop a search toolkit from an English Wiktionary dump.
    \item For training and evaluation, we construct a new dataset for neologism-aware machine translation and develop an accompanying search toolkit derived from an English Wiktionary dump.
\end{itemize}

% \begin{itemize}
%     \item Construction of a new multilingual neologism machine translation dataset named Neko. It contains solid neologism definitions and translation pairs covering 16 languages, mitigating the data shortage of neologism machine translation.
%     \item To the best of our knowledge, our research represents the pioneering effort in exploring the effect of agentic reinforcement learning based on GRPO on neologism machine translation. 
%     Furthermore, we propose an new RL approach based on GRPO that incorporates an adaptive sampling method according to translation difficulty.
%     \item  Our model \modelname{}, which actively retrieves external dictionary items to refine translation of neologisms in a multi-turn way when conducting reasoning, achieves the state-of-the-art performance, outperforming traditional SFT and RAG systems and other MT-specific models on neologism machine translation.
% \end{itemize}

\section{Overcoming Data Scarcity for Neologism-Aware Machine Translation}

\subsection{Construction of Neko Dataset}
\begin{figure*}[ht]
    \centering
    \includegraphics[width=0.85\textwidth]{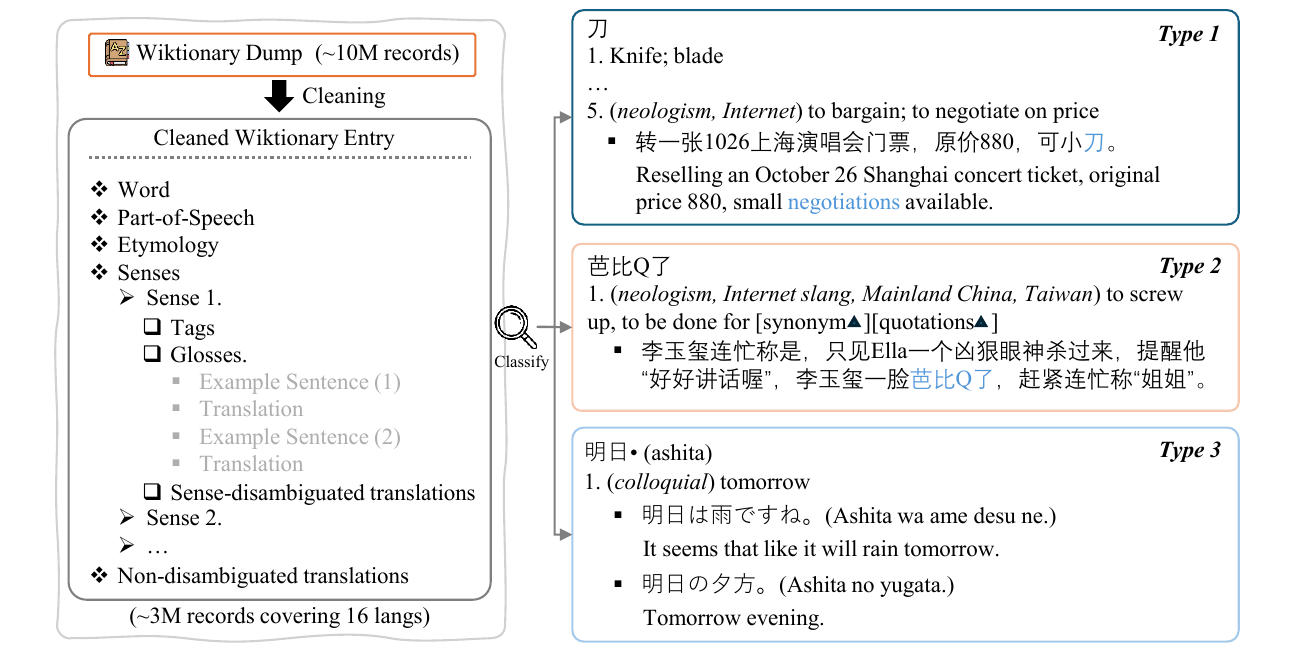}
    \caption{The construction process of our \datasetname{} benchmark. We clean about 10M records from a Wiktionary dump. In each cleaned word entry, we have word, part-of-speech, etymology, senses and non-disambiguated translations. In word senses, we have tags, such as ``neologism'', ``Internet'', glosses and so on. Glosses contain definitions of words. Example sentences and translations are available for some words. We obtain three types of Wiktionary word entries from the cleaned records. Blue words are neologisms or their translations.}
    \label{fig:example_neko}
\end{figure*}

% \subsection{}
\label{method_neko}

\begin{figure}[t]
    \centering
    \begin{subfigure}{0.4\linewidth}
        \centering
        \includegraphics[width=\linewidth]{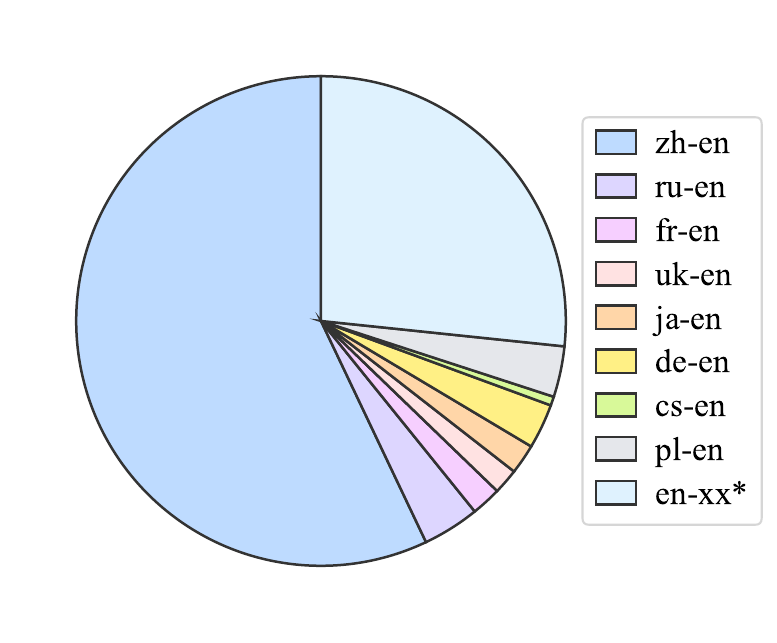}
        \caption{Proportions of translation pairs by language pair in the test split of the \datasetname{} dataset.}
        \label{fig:proportion_neko_test}
    \end{subfigure}
    \hfill
    \begin{subfigure}{0.48\linewidth}
        \centering
        \includegraphics[width=\linewidth]{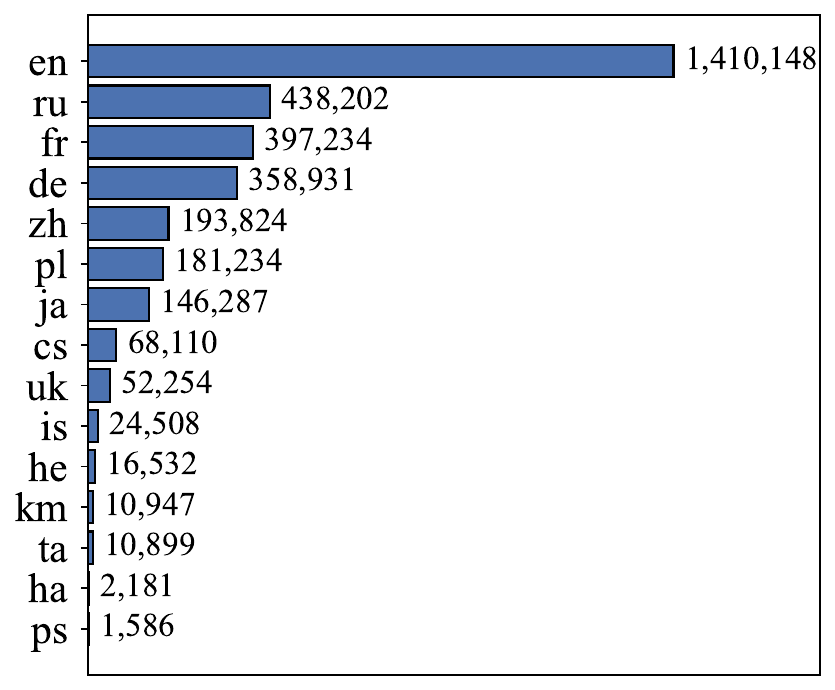}
        \caption{Number of word definitions per language in the dictionary constructed for the search toolkit.}
        \label{fig:distribution_lang_dict}
    \end{subfigure}
    \caption{Statistics of the test split of \datasetname{} dataset and the constructed dictionary in the search toolkit.}
\end{figure}

The first challenging issue in neologism-aware machine translation is that there is no solid testbed which contains high-quality translation pairs and neologism definitions.
Although the previous study Neo-bench~\citep{zheng-etal-2024-neo} has a neologism-aware machine translation subset, it is relatively small (240 in total) and contains only one language pair and not publicly available as shown in Table~\ref{tab:data-comparison}.

Wiktionary\footnote{\url{https://www.wiktionary.org/}} is a free and multilingual project that hosts a large number of word entries containing annotated tags including ``neologism'', word definitions and optional example sentences and their translations.
Therefore, we select the English Wiktionary as our data source to build our new dataset, \datasetname{}, which is short for \textbf{N}eologism \textbf{E}xample Sentences from Wi\textbf{K}ti\textbf{O}nary. Note that the English Wiktionary is a multilingual dictionary in which glosses are written in English.
Sixteen languages spanning across WMT20-23 are chosen as our research languages. The sixteen languages are English (en), Icelandic (is), Russian (ru), Khmer (km), Hausa (ha), German (de), Japanese (ja), Chinese (zh), Czech (cs), Ukrainian (uk), Tamil (ta), Polish (pl), French (fr), Hebrew (he), Croatian (hr) and Pashto (ps).

\begin{table}[ht]
\centering
\resizebox{0.45\textwidth}{!}{%
\begin{tabular}{@{}lccc@{}}
\toprule
Dataset                    & \# Langs &  \# Examples (train/val/test)  \\ \midrule
Neo-bench                  & 2        & 240 (~-~/~-~/~240~)            \\
Neko (ours)                & 16       & 11,521 (~10,425~/~83~/~1,013~)         \\ \bottomrule
\end{tabular}%
}
\caption{Comparison of neologism-aware machine translation datasets. The number of translation pairs in training, validation and testing splits of \datasetname{} are listed in Table~\ref{tab:neko_statistics_train} and~\ref{tab:neko_statistics_val_test}. The test split of the \datasetname{} dataset has 743 examples for other-language-to-English directions and 270 examples for English-to-other-language directions in total.}
\label{tab:data-comparison}
\end{table}

We cleaned around 10M records from a recent English Wiktionary dump\footnote{Specifically, we use the raw data from~\url{https://kaikki.org/dictionary/rawdata.html} (JSONL) dated 2025-08-23.} and obtained 3,312,877 records in 16 languages in which 3,606 word entries are tagged as neologisms.
We then classify the cleaned records into three types:

\begin{itemize}
    \item \textbf{\textit{Type 1}} represents the word entries that are neologisms and have example sentences and corresponding translations. 
    \item \textbf{\textit{Type 2}} denotes the word entries that are neologisms and have example sentences without translations.
    \item \textbf{\textit{Type 3}} comprises all word entries except \textbf{\textit{Type 1}} and \textbf{\textit{Type 2}}. Note that this type of data may contain neologisms without example sentences and corresponding translations, besides regular word entries.
\end{itemize}

The entire cleaning and processing pipeline is illustrated in Figure~\ref{fig:example_neko}.
After data cleaning and classification, we obtained 831 \textbf{\textit{Type 1}}, 2,581 \textbf{\textit{Type 2}} and 3,309,465 \textbf{\textit{Type 3}} word entries, respectively.

\paragraph{Validation and test data.} We extract the example sentences and their translations from \textbf{\textit{Type 1}} word entries. After that, we split them into the validation and test sets because these translation pairs have been carefully examined by Wiktionary editors and thus they are of high-quality and human-verified. 
This original test split only contains other-lang-to-English translation directions since we use the English Wiktionary. Other language versions of Wiktionary were also examined (e.g., the Japanese and Chinese Wiktionary dumps); however, only a very limited number of English-to–other-language translation pairs containing neologisms were found. Therefore, we selected 270 word entries from \textbf{Type 2} word entries containing neologism definitions and example sentences and used them for reference-free evaluation. 
The proportions of language pairs in the test split can be found in Figure~\ref{fig:proportion_neko_test}. En-xx* in the figure indicates that this translation direction is evaluated in a reference-free way provided with the neologism definitions. 

\paragraph{Training data.} For training data, we extract 700 example sentences with neologism definitions from \textbf{\textit{Type 2}} data except those used in the reference-free evaluation and use \gptfivename{} to translate the source example sentences into the other 15 languages with the neologism definitions in the prompt. 
% The prompt used for translation can be found in Appendix~\ref{app:neko_cons_details}.
In addition, we also obtain the aligned spans $\mathcal{S}$ in the target sentences for the neologisms in the source sentences. 
The details of the alignment process and the human validation on the translation quality can be found in Appendix~\ref{app:neko_cons_details} and~\ref{app:validation_type_2} respectively.
The final step is to convert the translation pairs into the correct prompt formats for training and evaluation.

\subsection{Search Toolkit}
\label{search_setting}
\paragraph{Dictionary.}
% To control the reproducibility of experiments, we use the definitions of all word entries from our cleaned records as the retrieval corpus. 
Unlike question answering (QA) and mathematical tasks, an MT-oriented dictionary is required for translation agents.
Specifically, we collect word definitions from our cleaned 3,312,877 word entries.
Each word definition includes ``part-of-speech'', ``etymology'', ``senses'', and optional ``word translations'' in other languages.
These word definitions are compiled into a dictionary for later retrieval by translation agents.
% The distribution of number of word definitions in the constructed dictionary is shown in Figure~\ref{fig:distribution_lang_dict}.
The distribution of word definition counts for different languages in the constructed dictionary is presented in Figure~\ref{fig:distribution_lang_dict}.
To better capture word semantics, we encode the entire word definition including “part-of-speech”, “etymology”, “senses”, and optional “word translations” into a semantic vector. This is because some headwords share the same form but differ in their glosses (senses) and etymologies; encoding headwords alone would be insufficient to distinguish these distinct lexical entries.

\paragraph{Retrieval Engine.}

Multilingual embedding models~\citep{sturua2024jina,zhao-etal-2024-leveraging,chen-etal-2024-m3,miao-etal-2024-enhancing,zhang2025qwen3} have demonstrated superior performance across a variety of retrieval tasks.
We utilize the widely used multilingual embedding model~\texttt{bge-m3}\footnote{\url{https://huggingface.co/BAAI/bge-m3}}~\citep{chen-etal-2024-m3} to obtain dense representations of items in the constructed dictionary.
The reason why we choose \texttt{bge-m3} is that it is one of SOTA embedding models for retrieval and the pretraining data of \texttt{bge-m3} covering 194 languages contains a large amount of translation-related corpora, for example, NLLB~\citep{nllbteam2022languageleftbehindscaling}, CCMatrix~\citep{schwenk-etal-2021-ccmatrix}, which should reduce the possible bias or domain mismatch. 
% For sanity check, we also analyzed the retrieval performance of \texttt{bge-m3} by human analysis in \S~\ref{sec_human_analysis} after training. 
% 76.71\% of retrieved results given related neologism queries contain the corresponding neologism definitions. This demonstrates the effectiveness of \texttt{bgm-m3} in our setting.
We use cosine similarity to measure the relevance of two items.
The faiss library\footnote{\url{https://github.com/facebookresearch/faiss}}~\citep{johnson2019billion,douze2024faiss} is leveraged to retrieve relevant items given a query.

The search toolkit that we proposed based on the Wiktionary is quite flexible.
You can add custom words or replace ambiguous ones at any time to further refine the dictionary knowledge base. This framework can also be used with other dictionaries in many different languages because we use a multilingual embedding model,~\texttt{bge-m3}.
The \texttt{bge-m3} can also be replaced with the latest multilingual embedding models, MT-specific or dictionary-specific embedding models.

\section{RL Algorithm}

The proposed approach comprises two key components: a carefully designed reward scheme and a translation difficulty-based adaptive sampling for GRPO~\citep{shao2024deepseekmath}.
We begin by establishing the notation adopted throughout this section.
% Let $x$ denote the source text, and $\hat{y}$ the model-generated translation.
% We use $y^\subtxt{ref}$ to represent the reference translation.
Let $x$, $\hat{y}$, and $y^\subtxt{ref}$ denote the source text, the model-generated translation, and the reference translation, respectively.

\paragraph{Prompt Template.}
To train our models, we design a prompt template that guides the initial LLM through a series of predefined steps for translating the source texts. 
The minimal output structure comprises four main steps: an initial reasoning process, a search process, a subsequent reasoning process, and a final translation generation step. The reasoning and search processes can be iteratively repeated by the LLM as needed.
The predefined steps are not compulsory. 
% We add a reward to encourage the model to generate the correct final translation format.
The prompt template can be found in Table~\ref{tab:training_template_sim}.

\begin{table}[h]
\centering
\resizebox{0.45\textwidth}{!}{%
    \begin{tabular}{p{12cm}}
        \toprule
        % \modelname{} Training Prompt                                        \\ \midrule
        You are a professional \textblock{\{src\_lang\}} to \textblock{\{tgt\_lang\}} translator. Please translate the following text from \textblock{\{src\_lang\}} to \textblock{\{tgt\_lang\}}. \\
        You must conduct reasoning inside \thinkblock{<think>} and \thinkblock{</think>} first. 
        After reasoning, you can use the search tool by enclosing your query within \searchblock{<search>} and \searchblock{</search>}. 
        The query could be unfamiliar terms, relevant keywords, or example sentences. 
        The search tool will then return the top results between \infoblock{<information>} and \infoblock{</information>}. 
        You can use the returned information to improve your translation. 
        You can reason and search as many times as you want. 
        In the end, you should directly provide the final translation inside \translationblock{<translation>} and \translationblock{</translation>} with a new line, without detailed illustrations.  \\
        ...\\
        \textblock{\{src\_lang\}} text: \textblock{\{src\_text\}} \\
        \bottomrule                                                     
    \end{tabular}%
}
\caption{Training template of \modelname{}. The full prompt can be found in Table~\ref{tab:training_template} in Appendix~\ref{app:hyperparameter-prompt-templates}.}
\label{tab:training_template_sim}
\end{table}

\subsection{Reward Design}
\paragraph{Outcome reward.}
The outcome reward incorporates three components: a neologism-specific reward, a neural model reward, and a format reward.

We propose a neologism-specific reward $R_\subtxt{neo}$ to encourage accurate translation of neologisms. This reward is adapted from the terminology success rate metric used in the WMT Terminology Translation Track~\citep{semenov-etal-2023-findings}.
We lemmatize the generated translation $\hat{y}$ to obtain $\hat{y}_l$, along with the corresponding spans of neologisms in the target language. The reward is then computed by checking whether the lemmatized neologism spans exist in $\hat{y}_l$. 
The detailed computation procedure for $R_\subtxt{neo}$ is presented in Algorithm~\ref{algo:neo_reward}.
The neural model reward $R_\subtxt{neural}$ is designed to ensure semantic fidelity between $x$ and $\hat{y}$. 
It combines scores from XCOMET-XL~\citep{guerreiro-etal-2024-xcomet} and CometKiwi-DA-XL~\citep{rei-etal-2023-scaling} as follows:
\begin{equation}
    R_\subtxt{neural} = \delta \cdot s_\subtxt{XCOMET} + (1 - \delta) \cdot s_\subtxt{CometKiwi-DA},
\end{equation}
where $s_\subtxt{XCOMET}$, $s_\subtxt{CometKiwi-DA}$ represents the scores of $\text{XCOMET-XL}(x, y^\subtxt{ref}, \hat{y})$ and $\text{CometKiwi-DA-XL}(x, \hat{y})$ respectively, and $\delta$ is set to $0.5$.
We also incorporate a format indicator reward $\mathds{1}_{\{\subtxt{format}\}}$ to encourage the model to produce translations in the expected format. 
% In this translation format reward, we do not add constraints to reasoning part and only add a constraint for the final translation format. 
% The total reward is set to zero if the model fails to generate the final translation within the <translation> and </translation> tags.
Specifically, the total reward is set to zero if the model fails to enclose the final translation within the \texttt{<translation>} and \texttt{</translation>} tags.
The final reward can be denoted as follows if we use outcome-based reward for training:
\begin{equation}
    R = \mathds{1}_{\{\subtxt{format}\}} \cdot (\lambda R_\subtxt{neo} +  (1 - \lambda) R_\subtxt{neural}).
\end{equation}

\paragraph{Process reward.} 
% A process reward is also explored to check its effectiveness. 
% This reward provides a score $R_\subtxt{q}$ about the search queries based on whether the queries that the model generates during reasoning are related to the neologism.
We also explore a process reward to evaluate its effectiveness. This reward assigns a score $R_\subtxt{q}$ to the search queries based on whether the lemmatized queries generated by the model during reasoning contain the lemmatized neologism span.
The final reward can be denoted as follows if we incorporate the process reward:
\begin{equation}
    R = \mathds{1}_{\{\subtxt{format}\}} \cdot (\lambda R_\subtxt{neo} + \sigma R_\subtxt{q} +  (1 - \lambda - \sigma) R_\subtxt{neural}),
\end{equation}
where $\lambda$ and $\sigma$ are set to $0.1$ by default.

\subsection{Adaptive Sampling based on Translation Difficulty}
% \paragraph{Relative Quality Estimation Rollout (RQE).}

The source texts in the \datasetname{} dataset exhibit varying levels of difficulty. 
% In some cases, LLMs translate easier source texts into translations that are even better than human translations.
Difficult source texts require LLMs to engage in intensive reasoning and explore a wider range of trajectories prior to translation, resulting in a comparatively slower learning process.
To effectively capture the distinction, we define a metric ``translation difficulty''  based on relative quality estimation scores (RQE) to measure how hard it is to translate source texts.
% The rewards in math and coding tasks are usually defined as 0/1 (failed/passed) in recent RL approaches.
% Unlike these tasks, 
The definition is as follows: 
\begin{definition}[\textit{Translation Difficulty}]\label{def: translation_challenge_degree}
    % \begin{equation}
    %     r_i = \alpha(t) (\Phi(x_i, y_i^\mathrm{ref}) - \Phi(x_i, \hat{y_i})) + \beta,
    % \end{equation}
        
    \begin{equation}
        v = \Phi(x, y^\subtxt{ref}) - \Phi(x, \hat{y}),
    \end{equation}
\end{definition}
where $\Phi$ denotes the quality estimation model and $\Phi(x, y) \in [0, 1]$.
The quality estimation model $\Phi$ is CometKiwi-DA-XL.
% Quality estimation models measure the quality of a translation by a float value (from 0.0 to 1.0).  This value provides more information than 0/1 rewards in math and coding tasks in which we need multiple inferences to obtain a relative value reflecting the ``difficulty'' which largely increases the computational cost~\citep{li2025know}.
We adopt this value rather than $1 -  \text{XCOMET-XL} (x, y^\subtxt{ref}, \hat{y})$ because LLMs can, in some cases, produce translations that surpass human references, a scenario that $1 - \text{XCOMET-XL}(x, y^\subtxt{ref}, \hat{y})$ fails to capture accurately.
% Translation difficulty indicates how hard it is to translate the source text by using the difference between two values $\Phi(x, y^\subtxt{ref})$ and $\Phi(x, \hat{y})$ from quality estimation models.
% $v > 0$ denotes that the quality of reference translation is better. $v < 0$ represents that the translation generated by current model in training is better, compared with reference translation. 
% Through translation difficulty, we encourage our model to generate more rollouts to increase the opportunity of obtaining better translations if the source text is challenging. On the other hand, we relatively reduce the number of rollouts to save the budget if the translation is easy.
By leveraging translation difficulty, we encourage the model to generate more rollouts for challenging source texts, thereby increasing the likelihood of obtaining higher-quality translations. Conversely, we reduce the number of rollouts for easier translations to conserve computational resources.

% After obtaining the translation difficulty $v$, we apply different coefficients according to the sign of $v$ to get $r$. Afterwards, we use $r$ to dynamically generate rollouts based on the translation difficulty.

Given the total budget of a batch $\mathcal{N} = |\mathcal{B}| * G$ where $\mathcal{B}$ is the input batch and $G$ is the group size in GRPO, 
we sample more trajectories for the training examples whose $v > 0$ during the rollout generation stage.
We sample fewer trajectories when $v < 0$.
The number of trajectories for one training example is $g = g_{\subtxt{initial}} \exp(r)$.
The computation of $r$ can be denoted as follows:
\begin{equation}
    r = 
    \begin{cases}
        \alpha v + \psi, v \ge 0 \\
        \gamma v + \psi, v < 0.
    \end{cases}
\end{equation}
We clamp $g$ between a minimum value $g_\subtxt{min}$ and a maximum value $G$ to prevent extreme values.
At last, if the total number of current rollouts $\sum_{j=1}^{j=|\mathcal{B}|} g_j < \mathcal{N}$, we assign the remaining budget $ \mathcal{N} - \sum_{j=1}^{j=|\mathcal{B}|} g_j $ to samples where $v_j > 0$ according to the ratio: $v_j / \sum_{j=1}^{j=|\mathcal{B}|} v_j$ where $v_j > 0$.

\paragraph{Policy Optimization.}

% GRPO~\citep{shao2024deepseekmath} obviates the need of value function approximation in PPO~\citep{schulman2017proximal} and uses the average reward of multiple sampled outputs, in response to the same user query, as the baseline. These modifications have enhanced the usability of GRPO and facilitated its widespread adoption across various domains.
Our policy optimization approach is based on GRPO.
In our RQE-based policy optimization, for each source text $x_i$, we sample multiple outputs $\{o_1, o_2, \dots, o_{g}\}$ according to the adaptive group size $g$. The $g$ is obtained through our RQE-based adaptive sampling method. The policy model $\pi_{\theta}$ is optimized by maximizing the following objective:

\begin{flalign}
\label{eq:grpo}
\begin{split}
&\mathcal{J}(\theta) = 
\mathbb{E}_{x \sim \mathcal{D}, \{ o_i \}_{i=1}^{g} \sim \pi_{\text{old}}( \cdot| x; \se)} \\
&\Bigg[\frac{1}{g} \sum_{i=1}^{g} \frac{1}{\sum_{t=1}^{|o_i|}  I(o_{i,t})} \sum_{t=1: I(o_{i,t})=1}^{|o_i|} \\
&\min \Bigg( \frac{\pi_{\theta}(o_{i,t} | x, o_{i,<t}; \se)}{\pi_{\text{old}}(o_{i,t} | x, o_{i,<t}; \se)} \hat{A}_{i,t}, \\
&\text{clip}\Bigg( \frac{\pi_{\theta}(o_{i,t} | x, o_{i,<t}; \se)}{\pi_{\text{old}}(o_{i,t} | x, o_{i,<t}; \se)}, 1 - \epsilon, 1 + \epsilon \Bigg) \hat{A}_{i,t} 
\Bigg)\\
&- \beta \mathbb{D}_{KL} \left[ \pi_{\theta} || \pi_{\text{ref}} \right]
\Bigg],
\end{split}
\end{flalign}
% where $g$ is the group size computed by our RQE approach. 
where $\pi_{\theta}$ and $\pi_{\text{ref}}$ are the current and reference policy models. $\pi_{\text{old}}$ is the old policy model used to generate rollouts.
$\se$ represents the retrieval engine. $\hat{A}_{i,t}$ is the advantage computed by using the relative rewards of outputs within the group $\{o_1, o_2, \dots, o_{g}\}$.
$\epsilon$ and $\beta$ are hyperparameters.
$I(o_{i,t})$ denotes the loss masking function. 
Following previous studies, like Search-R1~\citep{jin2025searchr} and ARPO~\citep{dong2025agentic}, we exclude the retrieved information in the policy optimization step by loss masking on the retrieved results.
Therefore,
$I(o_{i,t})$ is 1 when $o_{i, t}$ is a model-generated token and $I(o_{i,t})$ is 0 when $o_{i, t}$ is a retrieved token.

% \subsection{Search Strategy}
% \paragraph{Searching Always Succeed}
% \paragraph{Searching Always Failed}
% \paragraph{Searching in real-world}

\section{Experiments}

We address the following main research questions through experiments.
\paragraph{RQ1.} How to better address neologism-aware machine translation: using SFT, RL with reasoning, or teaching the model to consult dictionaries in the reasoning process using RL?
\paragraph{RQ2.} How should rewards be designed to enhance the performance of the neologism-aware machine translation agent?

\begin{table*}[ht]
\centering
% \small
\setlength{\tabcolsep}{4pt}
\renewcommand{\arraystretch}{1}
\resizebox{0.85\textwidth}{!}{%
    \begin{tabular}{l*{8}{c}}
        \toprule
        \multirow{2.5}{*}{\sc Model} & \multicolumn{4}{c}{\sc Neologism-Specific} & & \multicolumn{3}{c}{\sc Overall Quality} \\
        \cmidrule(lr){2-5} \cmidrule(lr){6-9}
        & \textsc{EXACT} & \textsc{FUZZY} & \textsc{LEM-EXACT} &  \textsc{LEM-FUZZY}  &  & \textsc{GEMBA(GPT5)} & \textsc{LJ(GPT5)} \\ 
        \midrule
        \multicolumn{9}{@{}l}{\textcolor{lightgray}{\textit{General Purpose LLMs}}} \\
        {Llama-3.1-8B-Instruct}  & 11.44 & 13.46 & 13.19 & 14.54 &  & 60.64  & 47.79 \\
        {Aya-23-8B} & 12.92 & 15.88 & 15.48 & 17.36  &  & 63.77  & 50.26             \\
        {Aya-expanse-8B} & 14.54 & 17.22 & 17.63 & 19.38  &  & 70.20  & 56.18       \\
        {Qwen3-4B} & 13.19 & 16.02 & 17.23 & 18.57 & & 65.24  & 51.94 \\
        {Qwen3-8B}  & 17.36 & 19.11 & 20.32 & 21.13  &  & 71.24  & 58.13 \\
        {Qwen3-14B} & 18.30 & 20.05 & 21.53 & 22.34  &  & 76.91  & 63.56 \\
        {Qwen3-32B} & 19.65 & 22.21 & 22.75 & 24.36  &  & 76.68  & 63.72 \\
        {GPT-OSS-20B}  & 18.84 & 22.07 & 21.67 & 23.42 &  & 73.96  & 62.62 \\
        \multicolumn{9}{@{}l}{\textcolor{lightgray}{\textit{Translation-Specific LLMs}}} \\
        {ALMA-R-7B} & 10.09 & 13.86 & 13.06 & 15.75  &  & 62.56  & 48.78 \\
        {MT-R1-Zero-7B-Mix}   & 16.29 & 18.30 & 18.84 & 19.11  &   & 67.22  & 53.80      \\
        {X-ALMA} & 13.46 & 15.07 & 16.55 & 17.09  &   & 71.28  & 58.87 \\
        {TowerInstruct-7B-v0.2}  & 11.71 & 14.40 & 14.27 & 16.15  &  & 60.20  & 47.49 \\
        {mExTrans-7B} & 13.46 & 18.71 & 16.42 & 19.78 &  & 70.37  & 57.42 \\
        {Hunyuan-MT-7B}  & 16.15 & 17.63 & 19.78 & 20.19 &  & 69.46  & 55.81 \\
        % \cdashlinelr{1-9}
        \midrule
        % \multicolumn{9}{@{}l}{\textcolor{lightgray}{\textit{Qwen3-4B}}} \\
        {Qwen3-4B} & 13.19 & 16.02 & 17.23 & 18.57 & & 65.24  & 51.94 \\
        {SFT-4B} & 13.73 & 16.96 & 17.23 & 18.84 & & 66.92 & 54.00 \\
        {GRPO-4B} & 13.73 & 15.74 & 16.82 & 18.98 & & 71.29  & 55.34 \\
        {\modelname{}-4B} & \underline{17.63} & \underline{19.11} & \underline{20.05} & \underline{21.53} & & \underline{72.93} & \underline{58.16} \\
        {\modelname{}-4B + process reward} & \textbf{19.25} & \textbf{23.55} & \textbf{23.55} & \textbf{27.19} & & \textbf{74.06} & \textbf{64.43} \\
        % \multicolumn{8}{c}{\textit{\textbf{Ours}}} \\
        % \multicolumn{9}{@{}l}{\textcolor{lightgray}{\textit{Qwen3-8B}}} \\
        \midrule
        {Qwen3-8B}  & 17.36 & 19.11 & 20.32 & 21.13  &  & 71.24  & 58.13 \\
        {SFT-8B}       & 15.61 & 18.17 & 19.11 & 20.32 &   & 72.39   & 60.01 \\

        % \multicolumn{10}{@{}l}{\textcolor{lightgray}{\textit{w/o. Search Reward and training with definitions}}} \\

        % GRPO wo search
        {GRPO-8B}      & 17.63 & \underline{21.13} & 20.19 & 22.75&  & 72.84 & 61.11 \\ 
        % {Ours (GRPO, definitions only)} & 20.19 & 23.15 & 23.01 & 25.17 & & 82.63 & 52.76 & 61.69   \\
        {\modelname{}-8B}   & \textbf{22.34} & \textbf{26.24} & \textbf{25.84} & \textbf{28.67}  &  & \underline{78.28}  & \underline{66.40}    \\ 
        {\modelname{}-8B + process reward} & \underline{18.57} & \underline{21.13} & \underline{23.82} & \underline{26.78} & & \textbf{79.50} & \textbf{67.58} \\
        % {Ours (GRPO, definitions$+$example sentences)} & 21.27 & 23.28 & 24.90 & 26.38 & & 82.51  \\
        % {Ours (RQE-GRPO, definitions$+$example sentences)}   & 23.82 & 26.24 & 27.05 & 28.94 & & 83.28 &   &     \\ 
        % {Ours (GRPO, oracle)}  & 36.47 & 38.49 & 41.18 & 41.72 & & 83.83 &        \\ 
        % {Ours (RQE-GRPO, oracle)}    & \textbf{44.41} & \textbf{46.57} & \textbf{48.72} & \textbf{49.80} & & \textbf{84.41} &        \\ 
        % \multicolumn{10}{@{}l}{\textcolor{lightgray}{\textit{w. Search Reward}}} \\
        % {Ours (w. rqe, definitions only)}   &  & &  &  & &  & &     \\ 
        \bottomrule
    \end{tabular}%
}
% \caption{Main results of various models on the test split of our \datasetname{} dataset for the other-language-to-English directions.
% GEMBA(GPT5) column provides the results of using the exact GEMBA-DA-Ref prompt~\citep{kocmi-federmann-2023-large} with \gptfivename{}.
% ``LJ(GPT5)'' is neologism-aware ``LLM-as-a-judge'' evaluation since we provide \emph{neologism definition} in the prompt to aid the backbone LLM to assess the translation quality following the GEMBA-style prompt.}
\caption{Main results of various models on the test split of our \datasetname{} dataset for other-language-to-English translation directions. The ``GEMBA(GPT5)'' column reports results obtained using the original GEMBA-DA-Ref prompt~\citep{kocmi-federmann-2023-large} with \gptfivename{}. ``LJ(GPT5)'' denotes our neologism-aware LLM-as-a-judge evaluation, in which we supply the \emph{neologism definition} within a GEMBA-style prompt to help the backbone LLM better assess translation quality.}

\label{tab:main-result}
\end{table*}

\subsection{Setup}
In this section, we illustrate our experimental settings, including datasets and evaluation.
As described in \S~\ref{method_neko}, we split the example sentences and their translations from \textbf{\textit{Type 1}} of the \datasetname{} dataset into validation and test sets.
We utilize 700 example sentences from \textbf{\textit{Type 2}} of the \datasetname{} dataset and their translations as our train set.
The statistics about the train, val and test splits of \datasetname{} dataset are presented in Table~\ref{tab:data-comparison},~\ref{tab:neko_statistics_val_test} and~\ref{tab:neko_statistics_train} in Appendix~\ref{app:statistics}.
The performance of our model on the WMT24++~\citep{deutsch-etal-2025-wmt24} is also reported in Appendix~\ref{wmt24pp}. Baselines are introduced in Appendix~\ref{app:baseline}.
% The details of the train, validation and test sets can be found in Section~\ref{method_neko} and Appendix~\ref{app:neko_cons_details}.

% \miao{artificial neologisms.}
% Neologisms in Wiktionary may be exposed in training data of LLMs. To analyze the 
% Refer to 実在しないエンティティや出来事に関する合成文書を用いたRAGベンチマークの構築

\subsection{Evaluation}
We evaluate translation quality from two perspectives: neologism-specific quality and overall quality.

\paragraph{Neologism-Specific Metrics.}
For neologism-specific metrics, we use four metrics, exact translation success rate $s_{\subtxt{exact}}$, fuzzy translation success rate $s_\subtxt{fuzzy}$, lemma-exact translation success rate $s_\subtxt{lem-exact}$, and lemma-fuzzy translation success rate $s_\subtxt{lem-fuzzy}$. These scores use different string matching approaches, such as lemmatization, exact string matching and fuzzy string matching\footnote{For lemmatization, we use stanza (1.10.1). For fuzzy string matching, we use fuzzywuzzy (0.18.0). 
These four metrics are adapted from the terminology success rate used in terminology translation track of WMT~\citep{semenov-etal-2023-findings,semenov-etal-2025-findings}. 
Our metric implementation is based on this code snippet provided by the WMT25 terminology translation track: \url{https://colab.research.google.com/drive/1b_M4tHnJxYT1JHZakjq44YJJIVsOP38h?usp=sharing}.}. 
The four metrics aim to check if the translations of neologisms can be found in the reference translations or not. 
These metrics can complement each other to ensure the stability and accuracy of the evaluation.
The detailed computation procedures of neologism-specific metrics are presented in Appendix~\ref{app:neologism-metrics}.

\paragraph{Overall Quality Metrics.}
For overall quality, we mainly use LLM-as-a-judge evaluation. 
In this evaluation, we use the GEMBA-DA prompt~\citep{kocmi-federmann-2023-large} with \gptfivename{} to evaluate the translations.
In addition to the GEMBA evaluation, we also provide additional neologism gloss information in the prompt to aid the backbone LLM to better judge the translation quality from both neologism and overall semantic-matching perspectives.
The detail of the LLM-as-a-judge prompts can be found in Table~\ref{tab:llm_as_a_judge_prompt} and Table~\ref{tab:llm_as_a_judge_gemba_prompt}, Appendix~\ref{app:eval_details}.

\subsection{Experimental Results}

We present main results in Table~\ref{tab:main-result}. The results show that our training framework can improve the model performance and demonstrate the effectiveness of the reward design.

\paragraph{The improvement of neologism translation quality using SFT is limited.}
For the neologism-specific scores, SFT result shows that the fine-tuned models cannot elicit the ability of translating neologisms well even if they are finetuned on the training data. This result supports the assumption that the base model itself has little inherent knowledge about neologisms.

\paragraph{RAG struggles to follow instruction/gets lost in retrieved word definitions.}
We use the same queries as our model, that is, \modelname{}-4B and \modelname{}-8B in RAG for retrieval to ensure a fair comparison and investigate the effect of queries.
The results about neologism-specific scores and overall quality are display in Table~\ref{tab:rag_neologism_specific}, Appendix~\ref{app:rag_neologism_specific} and Table~\ref{tab:rag_overall_quality} respectively.
RAG outperforms other baselines in the neologism-specific scores, suggesting that the retrieved word entries could benefit the neologism translation even under the RAG setting. This result also demonstrates that the queries generated by our model are beneficial for translating neologisms. However, the overall quality of RAG is not satisfying and we observe that augmented prompts with retrieved word entries cause the model not to follow the instruction and hallucinate; for example,  it continually generates fake word entries.

\begin{table}[ht]
    \centering
    \resizebox{0.38\textwidth}{!}{%
    \begin{tabular}{l*{2}{c}}
    \toprule
    \textsc{MODEL}   & \textsc{GEMBA(GPT5)} & \textsc{LJ(GPT5)} \\
    \midrule
        {Qwen3-4B + RAG}  &  65.30 & 52.88 \\
        \rowcolor[HTML]{EEEEEE} {\modelname{}-4B} & {72.93} & {58.16} \\
        {Qwen3-8B + RAG}  &  69.57  & 56.14 \\
        \rowcolor[HTML]{EEEEEE} {\modelname{}-8B} &  {78.28}  & {66.40}    \\  
    \bottomrule
    \end{tabular}}
    \caption{Overall quality scores about the RAG ablation study.}
    \label{tab:rag_overall_quality}
\end{table}

\paragraph{Performance on English-to-other-language directions.}
Table~\ref{tab:en2other_eval} presents the LLM-as-a-judge results of various finetuning approaches on the English-to-Japanese and English-to-Chinese translation directions.
The results demonstrate the effectiveness of our training framework on English-to-other-language directions.
% Note that the dictionary used here is the same dictionary for other-language-to-English direction derived from the English Wiktionary. The glosses for the English neologisms are provided in the source language, English.
Note that the dictionary used here is the same one employed for the other-language-to-English directions, derived from English Wiktionary. The glosses for English neologisms are provided in the source language, i.e., English.

\begin{table}[h]
\centering
\resizebox{0.42\textwidth}{!}{%
\begin{tabular}{l*{2}{c}}
\toprule
                  & LJ(GPT5, en-ja) & LJ (GPT5, en-zh)   \\ \midrule
% Llama3.1-8B-Inst & 51.96 & 53.77 \\
% Aya-23-8B        & 51.42 & 49.27 \\
% Aya-expanse-8B   & 55.76 & 55.47 \\
% Qwen3-8B         & 58.07 & 63.38 \\
% Qwen3-14B        & 63.02 & 68.87 \\
% Qwen3-32B        & 65.20 & 70.53 \\
% GPT-OSS-20B      & 68.97 & 68.11 \\
% Hunyuan-MT-7B    & 67.29 & 63.37 \\
% X-ALMA           & 57.40 & 54.70 \\
% \midrule
Qwen3-4B-SFT     & 51.22 & 65.74 \\
GRPO-4B    & 56.82 & 63.24 \\
\rowcolor[HTML]{EEEEEE} NeoAMT-4B  & 55.69 & 65.87 \\
\rowcolor[HTML]{EEEEEE} NeoAMT-4B + process reward  & 57.93 & 69.53 \\ 
\midrule
Qwen3-8B-SFT     & 60.27 & 69.86 \\
GRPO-8B    & 64.96 & 70.51 \\
\rowcolor[HTML]{EEEEEE} NeoAMT-8B  & 68.80 & 76.84 \\ 
\rowcolor[HTML]{EEEEEE} NeoAMT-8B + process reward & 68.36 & 78.27 \\
\bottomrule
\end{tabular}%
}
\caption{LLM-as-a-judge reference-free evaluation results of English-to-Japanese and English-to-Chinese directions.}
\label{tab:en2other_eval}
\end{table}

\section{Analysis}
In this section, we present a series of analyses and ablation studies based on the \modelname{}-8B model. 
The detailed analysis can be found in the Appendix~\ref{app:more_analysis}.
We also present the LLM-as-a-judge results including GEMBA-DA and neologism-aware evaluation by language pairs in Figure~\ref{fig:radar_plot_per_lang} and XCOMET and MetricX24 results in Table~\ref{tab:xcomet-metricx24-result}.

\subsection{Latency of Reasoning-Retrieval Loop}
The reasoning-retrieval loop introduces additional latency compared to standard end-to-end MT systems. 
To quantitatively assess this latency overhead, we measure the inference time of both systems.
% Table~\ref{tab:actual_inference_time} shows the actual inference time of a standard MT system and our agentic model, NeoAMT, with the same model size under the same environment.
Table~\ref{tab:actual_inference_time} presents the inference time of a standard MT system and our agentic model, NeoAMT, using the same model size under identical conditions.

\begin{table}[h]
    \centering
    \resizebox{0.42\textwidth}{!}{%
    \begin{tabular}{lcc}
        \toprule
         &  Standard MT & NeoAMT-8B \\
        \midrule
        Prompt template	& Table~\ref{tab:dt_prompt} & Table~\ref{tab:training_template} \\
        Total time (s) & $247.86$ & 	$571.37$ \\
        Avg. time / sent (s) & $0.33$ & $0.77$ \\
        Number of examples & $743$ & $743$ \\
        \bottomrule
    \end{tabular}
    }
    \caption{Comparison of actual inference time. The main component of the standard MT system is a Qwen3-8B model equipped with the direct translation prompt shown in Table~\ref{tab:dt_prompt}. The experiments are conducted on a computing node with dual AMD EPYC 7742 CPUs and 8 NVIDIA A100 GPUs, each with 80 GB of device memory. The detailed environment setup is shown in Table~\ref{tab:latency_testing_env}.}
    \label{tab:actual_inference_time}
\end{table}

% From the table, we can observe that the average generation time consumed by NeoAMT-8B for one example is about 0.77s, which is good enough to deploy it directly even without any optimization.
As shown in the table, the average generation time of NeoAMT-8B is approximately $0.77$ seconds per example, which is sufficiently efficient for direct deployment even without further optimization.
In addition, the reasoning-retrieval loop is conducted once in most cases as shown in Figure~\ref{fig:search_turn}. When our model thinks that it has the knowledge to deal with the source sentences, it will not retrieve any words from the external knowledge base.

% Leveraging hardware accelerators (GPUs) and software focusing on inference speed optimization (e.g., vllm and sglang) is a good option when deploying this system into applications considering the tradeoff between standard but static and frozen end-to-end MT systems and our flexible agentic MT approach.

\subsection{Impact of RQE}
We investigate the impact of RQE-based adaptive sampling in this section.
We train another model using the same base model and hyperparameters without RQE sampling, that is, \modelname{} w/o. RQE.
The results of \modelname{} and \modelname{} w/o. RQE are presented in Table~\ref{tab:rqe_neologism_specific} and Table~\ref{tab:rqe_overall_quality}.
RQE improves the model performance in almost all cases, especially for neologism-specific scores. 
% The most significant one is when the retrieved information contains example sentences and translations.
The results demonstrate the effectiveness of RQE-based policy optimization.

\begin{table}[h]
\centering
% \small
\setlength{\tabcolsep}{4pt}
\renewcommand{\arraystretch}{1}
\resizebox{0.42\textwidth}{!}{%
    \begin{tabular}{l*{4}{c}}
        \toprule

        \textsc{MODEL} & \textsc{EXACT} & \textsc{FUZZY} & \textsc{LEM-EXACT} &  \textsc{LEM-FUZZY}  \\ 
        \midrule
        \rowcolor[HTML]{EEEEEE} {\modelname{}}   & \textbf{22.34} & \textbf{26.24} & \textbf{25.84} & \textbf{28.67}      \\
        {~~~~~~w/o. RQE} & 20.19 & 23.15 & 23.01 & 25.17 \\
        \bottomrule
    \end{tabular}%
}
\caption{Neologism-specific scores about the ablation study of RQE.}
\label{tab:rqe_neologism_specific}
\end{table}

\begin{table}[h]
\centering
% \small
\setlength{\tabcolsep}{4pt}
\renewcommand{\arraystretch}{1}
\resizebox{0.4\textwidth}{!}{%
    \begin{tabular}{l*{3}{c}}
        \toprule

        \textsc{MODEL} & \textsc{GEMBA(GPT5)} & \textsc{LJ(120B)} & \textsc{LJ(GPT5)} \\ 
        \midrule
        \rowcolor[HTML]{EEEEEE} {\modelname{}}   & \textbf{78.28} & \textbf{63.72} & \textbf{66.40}    \\
        {~~~~~~w/o. RQE} &  77.65 & 61.69 & 64.50 \\
        \bottomrule
    \end{tabular}%
}
\caption{Overall quality scores about the ablation study of RQE.}
\label{tab:rqe_overall_quality}
\end{table}

\subsection{Human Evaluation}

We conduct a human evaluation to compare the translation quality of three models, Qwen3-8B, our base model, Hunyuan-MT-7B, a state-of-the-art MT model, and our model~\modelname{}. For each input sentence, three anonymized translation outputs (one per model) are presented to two human volunteers, who were asked to rank them from best to worst in terms of translation quality given the source sentence, reference translation and neologisms and their definitions.
% Two human volunteers are asked to compare three anonymized candidate translations. 
We sample 100 examples from the Chinese-English translation test split and we remove those examples in which two or more of the three translation candidates are identical. 
The result is shown in Table~\ref{tab:human-eval}.
This result verifies the result of LLM-as-a-judge evaluation and demonstrates the effectiveness of our training framework.
% The screenshot of the ranking annotation application can be found in Appendix~\ref{app:screenshot}. 
% We also provide some examples of translations generated by the three models in Table~\ref{tab:case-study}, Appendix~\ref{app:case_study}.

\begin{table}[h]
\centering
\resizebox{0.42\textwidth}{!}{%
\begin{tabular}{@{}lcccc@{}}
\toprule
Model         & \% 1st & \% 2nd & \% 3rd & Average Rank \\ \midrule
Qwen3-8B      &   3.33        &    25.56           &  71.11        &  2.68             \\
Hunyuan-MT-7B &  11.67        &    62.78           &   25.56           &  2.14            \\
\modelname{}  &  85.00        &    11.67           &    3.33           &   1.18           \\ \bottomrule
\end{tabular}%
}
\caption{Human evaluation.}
\label{tab:human-eval}
\end{table}

\subsection{Sanity Check about Thinking Paths of \modelname{}}
\label{sec_human_analysis}
We analyze the success and failure patterns by carefully investigating the thinking paths of our model~\modelname{} from three perspectives: (1) how many queries are related to neologisms? (2) do the retrieved results contain the neologism definitions? (3) does \modelname{} use the retrieved results for thinking?  
We randomly sample 100 model responses from the Chinese-English translation pairs.
Human volunteers are asked to check the model responses from the above three points.
The analysis shows that 86.90\% of search queries are related to neologisms. 76.71\% of retrieved results given related neologism queries contain the neologism definitions. 96.43\% of the model thinking paths after receiving the retrieved results use the retrieved neologism information for reasoning.
This indicates that (1) most of the model queries are related to neologisms; (2) the retrieval component still has room for improvement and is likely the dominant performance bottleneck; (3) \modelname{} actively utilizes the retrieved information in its reasoning paths.  

\subsection{Case Studies on Thinking Patterns}
\begin{figure}
    \centering
    \includegraphics[width=0.42\textwidth]{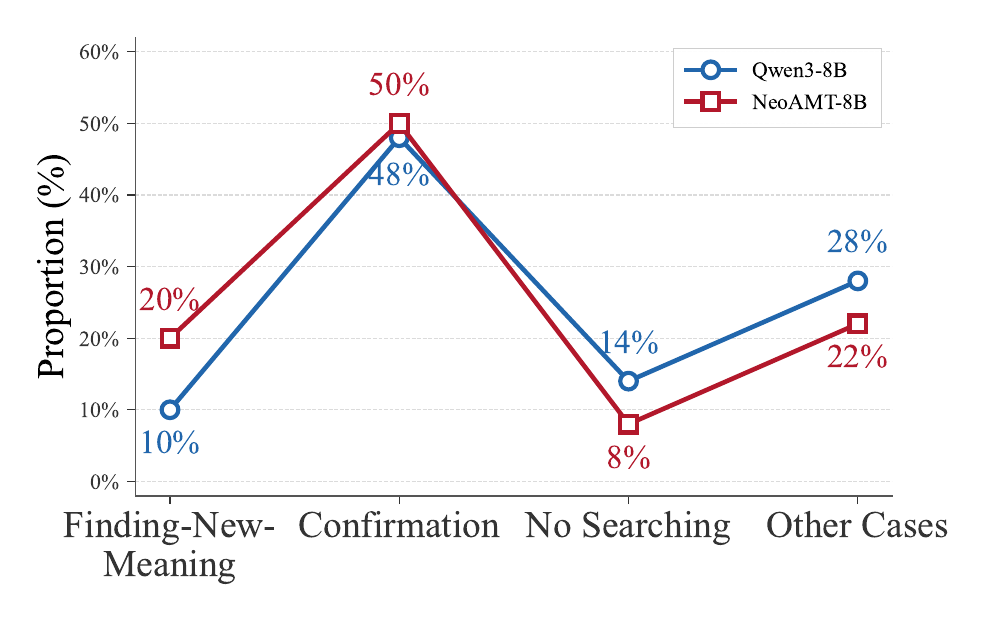}
    \caption{Thinking pattern distribution.}
    \label{fig:thinking_pattern_distribution}
\end{figure}

\begin{figure}
    \centering
    \includegraphics[width=0.42\textwidth]{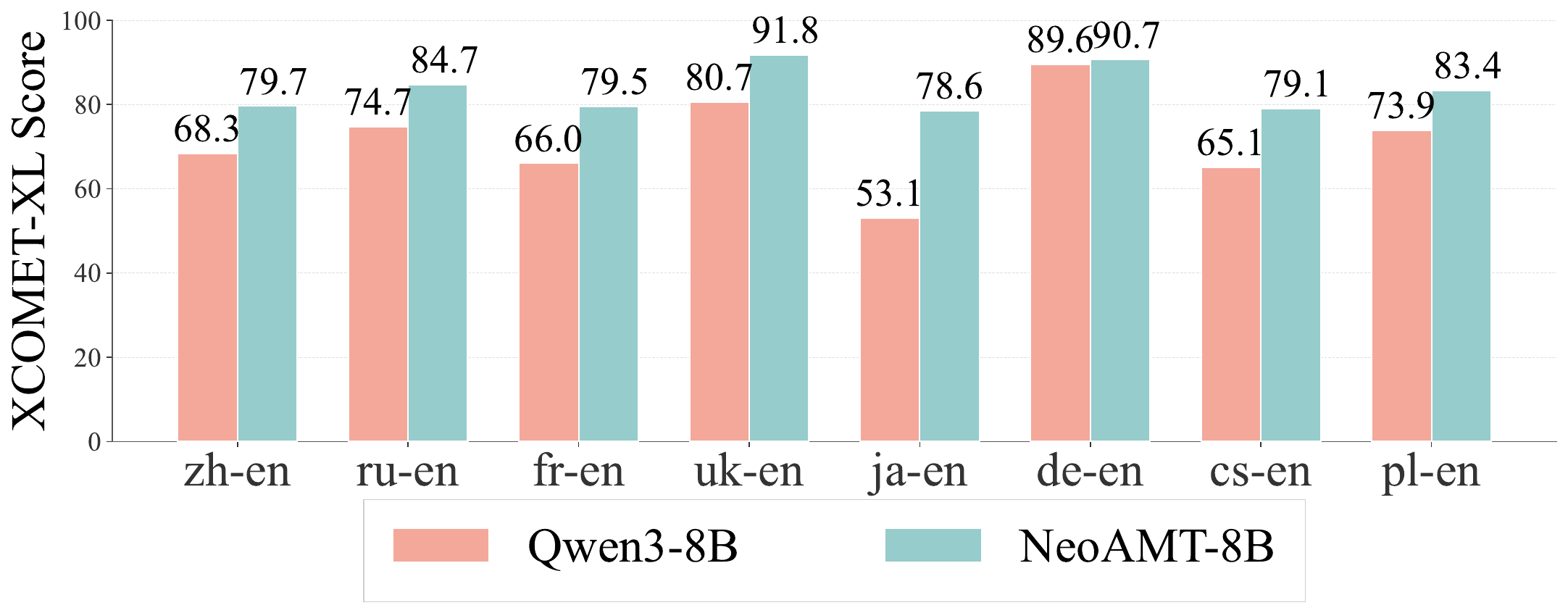}
        \caption{Comparison of XCOMET-XL scores.}
    \label{fig:comparison_xcomet_scores}
\end{figure}

We observe two intriguing thinking patterns in our model \modelname{} outputs:
\textbf{finding new meaning pattern} and \textbf{confirmation pattern}.
\textbf{Finding new meaning pattern} denotes the cases where \modelname{} first makes an incorrect translation or has no idea about neologism after the initial reasoning, but after searching and finding the new meaning of the query word by itself, the model corrects its reasoning path and generates the translation accordingly.
% In this pattern, the search function is used for finding new information.
\textbf{Confirmation pattern} indicates the cases in which \modelname{} generates candidate translations about neologisms in its reasoning process at first, and one of them is a correct assumption about the neologism. The model then uses the external search tool to verify the correct assumption and generates final translation. 
The full examples of two patterns can be found in Table~\ref{tab:find_new_meaning_example_our_model0} and Table~\ref{tab:confirm_pattern_example_model}, Appendix~\ref{app:ex} respectively. 
% We sample 100 Chinese-English translations from the base model, Qwen3-8B and NeoAMT-8B when they are configured using the same Wiktionary search toolkit and prompt to further investigate the change of the thinking pattern distribution.
To further investigate the shift in thinking pattern distribution, we sample 100 Chinese-English translations from the base model, Qwen3-8B, and NeoAMT-8B, all configured with the same Wiktionary search toolkit and prompt.
% The percentages of different thinking patterns are shown in Figure~\ref{fig:thinking_pattern_distribution}.
Figure~\ref{fig:thinking_pattern_distribution} illustrates the distribution of different thinking patterns.
The XCOMET-XL scores of the two models on the \datasetname{} test set are presented in Figure~\ref{fig:comparison_xcomet_scores}.
% We can observe that (1) the confirmation thinking pattern dominates for both models; (2) the RL training makes the trained model more actively use the search tool and the trained model can utilize the external dictionary more effectively from the table; (3) The translation quality of NeoAMT-8B is much higher compared with Qwen3-8B, which indicates that the training process makes the model excel at leveraging the external search tool to generate better translations.
We observe that: (1) the confirmation thinking pattern dominates in both models; (2) RL training encourages more active use of the search tool, enabling the trained model to leverage the external dictionary more effectively; (3) NeoAMT-8B achieves substantially higher translation quality than Qwen3-8B, indicating implicitly that the training process enhances the model's ability to leverage external search tools for improved translation.

\section{Related Work}
\paragraph{Improving Machine Translation via Reinforcement Learning.}
% Utilizing reinforcement learning to improve machine translation~\citep{wu-etal-2018-study,Choshen2020On,kiegeland-kreutzer-2021-revisiting} is an on-going research direction that provides a different way from the pretraining, supervised fine-tuning approaches~\citep{xu2024a,cui-etal-2025-multilingual}.
Recent studies on this direction mainly focus on how to design the reward empowered with test-time scaling. For example, MT-R1-Zero~\citep{feng-etal-2025-mt-r1} explores the effect of mixed rule-based and neural model-based metric score as rewards.
SSR-Zero~\citep{yang2025ssr} utilizes self-evaluation score as the reward for training.
TAT-R1~\citep{li2025tat} employs word alignment~\citep{jalili-sabet-etal-2020-simalign,nagata-etal-2020-supervised,wu-etal-2023-wspalign,latouche-etal-2024-binaryalign,miao-etal-2025-improving} to design rewards.
DeepTrans~\citep{wang2025deep} uses DeepSeek-V3~\citep{liu2024deepseek} to generate format, thought and translation rewards.
% All method mention above rely on the reasoning capability of LLMs elicited by internal frozen parametric knowledge. Different from these methods, our study explores how to enable LLMs to actively use external search tools in the reasoning paths to translation neologisms with reinforcement learning.

\paragraph{Neologism Translation.}
Neologism translation is a challenging task because of rapid change of languages and the insufficient data.
\citet{lerner-yvon-2025-towards} try to solve this issue using term definitions and in-context learning.
% Towards the Machine Translation of Scientific Neologisms
Neo-bench~\citep{zheng-etal-2024-neo} provides a benchmark dataset about neologisms in which one sub-task is machine translation. However, the machine translation subset in this benchmark is relatively small and not publicly available. 
% In addition, it contains one language pair. 
% Our \datasetname{} dataset contains much more neologism translation pairs and solid definitions about neologisms and covers 16 languages.
Other similar studies include social-media, culture-aware and poetry translations~\citep{yao-etal-2024-benchmarking,zhao2026benchmarkingmachinetranslationchinese,chen-etal-2025-benchmarking-llms}

\paragraph{Agentic Search.}

Recent work explores the integration of search tools into reasoning paths of LLMs to enhance accuracy and factuality on QA and math tasks~\citep{jin2025searchr,dong2025agentic,li-etal-2025-search}. For example, Search-R1~\citep{jin2025searchr} uses Wikipedia articles as the search source to improve the performance of models on QA tasks.
ARPO~\citep{dong2025agentic} utilizes the entropy to control the branching sampling during reasoning with external tools.
These works mainly focus on QA and math tasks. In this paper, we focus on an MT-specific domain, neologism-aware machine translation, which remains underexplored. 
% Furthermore, we explore the power of RL with the MT-specific search setting and introduce a new adaptive rollout approach based on translation difficulty to improve neologism machine translation.

% \kaiyan{增加几篇相关领域的比如文化词汇，文言文词汇类似的翻译工作。}

% \paragraph{Agentic LLMs.}

% Language Models that Think, Chat Better
% SeqPO-SiMT: Sequential Policy Optimization for Simultaneous Machine Translation
% Learning Optimal Policy for Simultaneous Machine Translation via Binary Search

\section{Conclusion}

In this paper, we first construct a new multilingual neologism-aware machine translation dataset named \datasetname{} that covers 16 languages and contains solid neologism definitions. 
The test split of this dataset provides a testbed for neologism-aware machine translation.
Furthermore, to fill in the gap that it remains unclear that whether reasoning-based RL with search tools improves neologism-aware machine translation, we conduct experiments to analyze the results of various methods and propose a new RL training framework, \modelname{}, consisting of a novel reward design and an adaptive sampling approach based on the translation difficulty to train the neologism-aware machine translation agent. 
The results of experiments demonstrate the effectiveness of our method.

\section*{Limitations}
We use a general-purpose emebdding model, \texttt{bge-m3}, for retrieving word entries in this paper. 
This model is not specifically designed for machine-translation related retrieval tasks though its training dataset contains translation-related corpora and the human analysis in \S~\ref{sec_human_analysis} also shows that there is room for improvement about the retrieval performance.
Another point is that \modelname{} does not follow the output format exactly specified in the user instruction prompt during the thinking process. The reason might be that our reward design only imposes a constraint on the final translation format.

\section*{Ethical Considerations}
This research aims to investigate the effect of agentic RL with search tools on the neologism-aware machine translation.
Our new dataset \datasetname{} is constructed from the English Wiktionary dump. Some of the word entries may be derogatory and related to politics and subcultures.
We process the data in compliance with the license of Wiktionary.
The views contained in the dataset are independent of the authors.
We have tried our best to remove such words.
In addition, we use various LLMs in this paper adhering to the respective model licenses.
We used ChatGPT and Claude Opus 4.6 to improve the writing of certain sections of this paper while preserving the original meaning.

% Bibliography entries for the entire Anthology, followed by custom entries
%\bibliography{anthology,custom}
% Custom bibliography entries only
\bibliography{custom}

\appendix

% \section{Example Appendix}
% \label{sec:appendix}

\clearpage

% \tableofcontents

\section{Implementation Details}
\label{app:implementation_details}
\subsection{Neologism-Specific Reward}
The computation procedure of the neologism reward is listed in Algorithm~\ref{algo:neo_reward}.

\begin{algorithm}[h]
\caption{Neologism-Specific Reward Computation}\label{algo:neo_reward}
\begin{algorithmic}[1]
\Require Source text $x$, generated translation $\hat{y}$, the set of corresponding spans of neologisms in $y^\subtxt{ref}$: $\mathcal{S}$, lemmatization function $\mathcal{L}$, exact string matching function $\mathcal{M}$.
\Ensure Neologism-specific reward $R_\subtxt{neo}$.

% \State Initialize lemma extractor: $\mathcal{L}$.
\State Initialize count: $c \gets 0$.
\State Initialize the number of spans in $\mathcal{S}$: $s = \lvert \mathcal{S} \rvert$.
\State Lemmatize $\hat{y}$ to obtain $\hat{y}_{l}$: $\hat{y}_{l} \gets \mathcal{L}(\hat{y})$.
\For{$s \in \mathcal{S}$} 
    \State Lemmatize $s$ to obtain $s_l$: $s_l \gets \mathcal{L}(s)$.
    \If{$M(s_l, \hat{y}_{l})$ is true }
        \State Increment count $c \gets c + 1$.
    \EndIf
\EndFor
\State \textbf{return} final neologism reward $R_\subtxt{neo} = c/s$.
\end{algorithmic}
\end{algorithm}

\subsection{Hyperparameters and Prompt Templates}
\label{app:hyperparameter-prompt-templates}
The full training template of our model \modelname{} is shown in Table~\ref{tab:training_template}.

Qwen3-4B and Qwen3-8B are base models for training. The code that we use is mainly adapted based on verl~\citep{10.1145/3689031.3696075}.

% The maximum number of tokens for user prompts is 1024.
% The maximum number of tokens for model responses (including retrieved results) is set to 4096.
% The maximum turns of searching of our agentic MT models is 3.
% The number of retrieved results is 5.

We use 8 NVIDIA A100 GPUs with 80GB memory to train our models and conduct inference. 
For software aspect, we use PyTorch (2.7.1)~\citep{10.1145/3620665.3640366} and our code is mainly adapted from the verl (0.5.0.dev0, GitHub commit id: 0e15c9b)~\citep{10.1145/3689031.3696075} and vllm~\citep{10.1145/3600006.3613165} (0.10.1.1) libraries.

\modelname{} models are trained for 1 epoch.
The ``train\_batch\_size'' is 32.
The ``ppo\_mini\_batch\_size'' is 16.
The ``ppo\_micro\_batch\_size\_per\_gpu'' is 2.
The maximum number of tokens for user prompts is 1024.
The maximum number of tokens for model responses (including retrieved results) is set to 4096.
The maximum turns of searching of our agentic MT models is 3.
The number of retrieved results is 5.
The maximum character length of retrieved result for a single turn is 2000.
The maximum training time for a single experiment is approximately 40 hours.
The $\lambda$ for $R_\subtxt{neo}$ is 0.1.
The $\alpha$, $\gamma$ and $\psi$ for RQE are 10, -5 and 0.0 respectively. 
The $g_\subtxt{min}$ and $G$ are 4 and 8, respectively. The $g_\subtxt{initial}$ is set to 4.
% More details can be found in Appendix~\ref{app:implementation_details}.

For the searching tool, we cache the retrieved results to speed up the training process. The caching mechanism is based on the ARPO implementation~\citep{dong2025agentic}.

\begin{table*}[ht]
\centering
\resizebox{0.85\textwidth}{!}{%
    \begin{tabular}{p{15cm}}
        \toprule
        % \modelname{} Training Prompt                                        \\ \midrule
        You are a professional \textblock{\{src\_lang\}} to \textblock{\{tgt\_lang\}} translator. Please translate the following text from \textblock{\{src\_lang\}} to \textblock{\{tgt\_lang\}}. \\
        You must conduct reasoning inside \thinkblock{<think>} and \thinkblock{</think>} first. 
        After reasoning, you can use the search tool by enclosing your query within \searchblock{<search>} and \searchblock{</search>}. 
        The query could be unfamiliar terms, relevant keywords, or example sentences. 
        The search tool will then return the top results between \infoblock{<information>} and \infoblock{</information>}. 
        You can use the returned information to improve your translation. 
        You can reason and search as many times as you want. 
        In the end, you should directly provide the final translation inside \translationblock{<translation>} and \translationblock{</translation>} with a new line, without detailed illustrations.  
        For example, \thinkblock{<think>} reasoning process here \thinkblock{<think/>} \searchblock{<search>} query here \searchblock{</search>}\infoblock{<information>} returned information \infoblock{</information>}\thinkblock{<think>} another reasoning process here \thinkblock{<think/>}\searchblock{<search>} another query here \searchblock{</search>}\infoblock{<information>} another returned information \infoblock{</information>}\thinkblock{<think>} yet another reasoning process\thinkblock{</think>}\translationblock{<translation>} final translation here \translationblock{</translation>}. \\
        \textblock{\{src\_lang\}} text: \textblock{\{src\_text\}} \\
        \bottomrule                                                     
    \end{tabular}%
}
\caption{Full training template of \modelname{}.}
\label{tab:training_template}
\end{table*}

We use the same hyperparameters as \modelname{} for the GRPO row except the prompt. The prompt we use for the GRPO row is presented in Table~\ref{tab:grpo_prompt}.

For SFT row in Table~\ref{tab:main-result}, the input and output formats that we use for training are shown in Table~\ref{tab:sft_prompt}. We use the ``fsdp\_sft\_trainer'' of verl. ``micro\_batch\_size\_per\_gpu'' is set to 4 and the total batch size is 32. The number of epoch is 3. The max length is set to 4096. We use the default setting of ``fsdp\_sft\_trainer'' for other hyperparameters.

\section{Latency Testing Environment}
\label{latency_testing_env}
The environment that we use for latency testing is the same as the training environment.
The details about the environment are shown as in Table~\ref{tab:latency_testing_env}.

\begin{table}[H]
\centering
\resizebox{0.45\textwidth}{!}{%
    \begin{tabular}{@{}ll@{}}
        \toprule
        % \textbf{Parameter} & \textbf{Value} \\
        % \midrule
        \multicolumn{2}{@{}l@{}}{\textit{Environment}} \\
        \midrule
        CPU & Dual AMD EPYC 7742 \\
        GPU & NVIDIA A100 80GB $\times$ 8 \\
        Software & vllm (0.10.1.1), verl (0.5.0.dev0, commit: 0e15c9b) \\
        Batch Size & 96 \\
        vLLM GPU Utilization & 0.8 \\
        Max Prompt Length & 1024 \\
        Max Response Length & 4096 \\
        Sampling & Greedy \\
        \midrule
        \multicolumn{2}{@{}l@{}}{\textit{Search Tool Setting}} \\
        \midrule
        Max Retry Count & 1 \\
        Timeout Limit & 10s \\
        Max Workers & 64 \\
        Max Search Turns & 3 \\
        \midrule
        \multicolumn{2}{@{}l@{}}{\textit{Retrieval Engine Setting}} \\
        \midrule
        Max Search Results & 5 \\
        Backend & FastAPI + FAISS \\
        Retriever Batch Size & 1 \\
        Loading \& Saving Cache & Enabled \\
        \bottomrule
    \end{tabular}
}
\caption{Details of the latency testing environment.}
\label{tab:latency_testing_env}
\end{table}

\section{More Evaluation Details}
\label{app:eval_details}
\subsection{Baselines}
\label{app:baseline}
We evaluate various models including general-purpose and MT-specific models on this dataset to report their performance on neologism-aware machine translation. 
We also present the results of Supervised Fine-Tuning (SFT), Retrieval-Augmented Generation (RAG)~\citep{NEURIPS2020_6b493230}, RL-based fine-tuning without searching (GRPO)~\citep{Guo2025-qv}. The reward for GRPO is  $R = \mathds{1}_{\{\subtxt{format}\}} *  R_\subtxt{neural}$.
We use the same queries as our model~\modelname{} for RAG to ensure a fair comparison.
For general-purpose models, we include 
Llama-3.1-8B-Instruct~\citep{grattafiori2024llama}, Aya-23-8B~\citep{aryabumi2024aya}, Aya-expanse-8B~\citep{dang2024aya}, Qwen3-8B, Qwen3-14B, Qwen3-32B~\citep{yang2025qwen3} and GPT-OSS-20B~\citep{agarwal2025gpt}.
We also conduct evaluation on several strong MT-specific models, including ALMA-R-7B~\citep{xu2024contrastive}, X-ALMA~\citep{xu2025xalma}, MT-R1-Zero-7B-Mix~\citep{feng-etal-2025-mt-r1}, TowerInstruct-7B-v0.2~\citep{alves2024tower}, Hunyuan-MT-7B~\citep{zheng2025hunyuan}, mExTrans-7B~\citep{wang2025extrans}.
The implementation details and prompts of various baselines, including general-purpose LLMs, MT-specific LLMs, RAG, SFT and GRPO, can be found in Appendix~\ref{app:implementation_details} and Appendix~\ref{app:eval_config}.

\subsection{Evaluation Configuration}
\label{app:eval_config}
We adopt the similar setting as MT-R1-Zero~\citep{feng-etal-2025-mt-r1} when evaluating the performance of general purpose baseline models on the test set.
For open weight models, we deploy them locally using the transformers implementation.
We use the sampling decoding strategy and the temperature is set to 0.2 and top\_p is 0.95. The maximum generation length is capped at 4096 tokens.
For MT-specific models, we use their corresponding prompt formats to obtain translations.
The detailed prompts can be found in Table~\ref{tab:alma_prompt} and Table~\ref{tab:hunyuan_mt_prompt}.

We mainly use the prompt format in Table~\ref{tab:dt_prompt} for general purpose models to obtain outputs because some of the general-purpose models tend to generate other content that is not directly related to translation. Therefore, we ask models to generate translations within the specific tags <translation> and </translation> to obtain the translations. We use the raw outputs as final translations when models fail to generate translations within the specific tags.
We apply their chat templates when necessary.
For XCOMET and MetricX-24, we use the XL version.

\begin{table}[h]
\centering
\resizebox{0.45\textwidth}{!}{%
    \begin{tabular}{p{8cm}}
        \toprule
        % Inference Prompt                                        \\ \midrule
        You are a professional \{src\_lang\} to \{tgt\_lang\} translator.
        Please translate the following text from \{src\_lang\} to \{tgt\_lang\}.
        Please provide the translation directly without reasoning 
        inside <translation> and </translation>. 
        For example, <translation> translation here </translation>. \\
        \{src\_lang\} text: \{src\_text\} \\
        \bottomrule                                                     
    \end{tabular}%
}
\caption{Direct translation prompt.}
\label{tab:dt_prompt}
\end{table}

\begin{table}[h]
\centering
\resizebox{0.45\textwidth}{!}{%
    \begin{tabular}{p{8cm}}
        \toprule
        % Inference Prompt                                        \\ \midrule
        Translate this from \{src\_lang\} into \{tgt\_lang\}: \\
        \{src\_lang\}: \{text\} \\
        \{tgt\_lang\}: \\
        \bottomrule                                                     
    \end{tabular}%
}
\caption{ALMA prompt~\citep{xu2024a,xu2024contrastive,xu2025xalma}.}
\label{tab:alma_prompt}
\end{table}

\begin{table}[h]
\centering
\resizebox{0.45\textwidth}{!}{%
    \begin{tabular}{p{8cm}}
        \toprule
        % Inference Prompt                                        \\ \midrule
Translate the following segment into \{tgt\_lang\}, without additional explanation. \\
\\
\{text\} \\
        \bottomrule                                                     
    \end{tabular}%
}
\caption{Hunyuan-MT prompt.}
\label{tab:hunyuan_mt_prompt}
\end{table}

For LLM-as-a-judge evaluation, the prompt with additional neologism definitions is presented in Table~\ref{tab:llm_as_a_judge_prompt}.
We also present the results of using the original GEMBA prompt in Table~\ref{tab:main-result}.
The original GEMBA prompt that we use is shown in Table~\ref{tab:llm_as_a_judge_gemba_prompt}.

\begin{table}[h]
\centering
\resizebox{0.45\textwidth}{!}{%
    \begin{tabular}{p{8cm}}
        \toprule
        % Inference Prompt                                        \\ \midrule
You are an expert in evaluating the quality of translations. \\
You will be given a source sentence, a reference translation, and a candidate translation. \\
The source sentence contains a neologism (a newly coined word or expression). \\
Your task is to determine how well the candidate translation captures the meaning of the source sentence, especially focusing on the neologism. \\
Please consider the following criteria when conducting your evaluation: \\
1. Neologism Quality (score: 0-50). \\
2. Overall Translation Quality (score: 0-50). \\
After evaluating the candidate translation based on the above criteria, please provide your assessment in the following format:
<evaluation> score </evaluation>. \\
The final ``score'' is a numerical value between 0 and 100. A higher score indicates a better translation.\\
Here is the information you will need for your evaluation: \\
Source Sentence: \{source\_sentence\} \\ 
Neologism and Its Meaning: \{neologism\} (\{neologism\_meaning\}) \\
Reference Translation: \{reference\_translation\} \\
Candidate Translation: \{candidate\_translation\} \\
        \bottomrule                                                     
    \end{tabular}%
}
\caption{Neologism-aware LLM-as-a-judge prompt. We provide additional neologism information in the prompt to help LLMs judge the translation quality from both neologism and overall perspectives following the GEMBA-style prompt~\citep{kocmi-federmann-2023-large}.}
\label{tab:llm_as_a_judge_prompt}
\end{table}

\begin{table}[h]
\centering
\resizebox{0.45\textwidth}{!}{%
    \begin{tabular}{p{8cm}}
        \toprule
        % Inference Prompt                                        \\ \midrule
Score the following translation from \{source\_lang\} to \{target\_lang\} with respect to human reference on a continuous scale 0 to 100 where score of zero means "no meaning preserved" and score of one hundred means "perfect meaning and grammar". \\ 
\\
\{source\_lang\} source: "\{source\_seg\}"\\
\{target\_lang\} human reference: \{reference\_seg\} \\
\{target\_lang\} machine translation: "\{target\_seg\}" \\ 
Score:  \\
        \bottomrule                                                     
    \end{tabular}%
}
\caption{GEMBA prompt.}
\label{tab:llm_as_a_judge_gemba_prompt}
\end{table}

\begin{table}[h]
\centering
\resizebox{0.45\textwidth}{!}{%
    \begin{tabular}{p{8cm}}
        \toprule
        % SFT Prompt                                        \\ \midrule
        \textbf{Input} You are a professional \{src\_lang\} to \{tgt\_lang\} translator. Please translate the following text from \{src\_lang\} to \{tgt\_lang\}. 
        Please provide the translation directly without reasoning inside <translation> and </translation>. For example, <translation> translation here </translation>. 
        \{src\_lang\} text: \{src\_text\} \\
        \textbf{Output} <translation> translation here <\/translation> \\
        \bottomrule                                                     
    \end{tabular}%
}
\caption{SFT prompt.}
\label{tab:sft_prompt}
\end{table}

For RAG row in Table~\ref{tab:main-result}, the prompt is shown in Table~\ref{tab:rag_prompt}.
The hyperparameters, such as temperature and top\_p, of the RAG baseline are the same as other baselines.

\begin{table}[h]
\centering
\resizebox{0.45\textwidth}{!}{%
    \begin{tabular}{p{8cm}}
        \toprule
        % Inference Prompt                                        \\ \midrule
You are a professional \{src\_lang\} to \{tgt\_lang\} translator. Please translate the following text from \{src\_lang\} to \{tgt\_lang\}. \\
Please provide the translation directly without reasoning inside <translation> and </translation>. For example, <translation> translation here </translation>. \\
\{src\_lang\} text: \{text\} \\ 
\{retrieved\_result\} \\
        \bottomrule                                                     
    \end{tabular}%
}
\caption{RAG prompt.}
\label{tab:rag_prompt}
\end{table}

\begin{table}[h]
\centering
\resizebox{0.45\textwidth}{!}{%
    \begin{tabular}{p{8cm}}
        \toprule
        % Inference Prompt                                        \\ 
        % \midrule
You are a professional \{src\_lang\} to \{tgt\_lang\} translator. Please translate the following text from \{src\_lang\} to \{tgt\_lang\}. \\
You must conduct reasoning inside <think> and </think> first.
In the end, you should directly provide the final translation inside <translation> and </translation> with a new line, without detailed illustrations. For example, <think> reasoning process here <think/><translation> final translation here </translation>. \\
\{src\_lang\} text: \{src\_text\} \\
        \bottomrule                                                     
    \end{tabular}%
}
\caption{GRPO prompt.}
\label{tab:grpo_prompt}
\end{table}

\subsection{Neologism-Specific Metrics}
\label{app:neologism-metrics}
The detailed procedure of computing neologism exact, fuzzy, lem-exact and lem-fuzzy scores are presented in Alogirhtm~\ref{algo:neo_regex_metric}, \ref{algo:neo_fuzzy_metric}, \ref{algo:neo_lem_regex_metric} and  \ref{algo:neo_lem_fuzzy_metric}.
These metrics complement each other to ensure stable evaluation of neologism translation and reduce evaluation bias.

\begin{algorithm}[h]
\caption{Neologism Exact Metric}\label{algo:neo_regex_metric}
\begin{algorithmic}[1]
\Require Source text $x$, generated translation $\hat{y}$, the set of corresponding spans of neologisms in $y^\subtxt{ref}$: $\mathcal{S}$, exact string matching function $\mathcal{M}$.
\Ensure Neologism exact score: \textsc{REGEX}.
\State Initialize count: $c \gets 0$.
\State Initialize the number of spans in $\mathcal{S}$: $s = \lvert \mathcal{S} \rvert$.
\For{$s \in \mathcal{S}$} 
    \If{$\mathcal{M}(s, \hat{y})$ is true}
        \State Increment count $c \gets c + 1$.
    \EndIf
\EndFor
\State \textbf{return} neologism exact matching score: $c/s$.
\end{algorithmic}
\end{algorithm}

\begin{algorithm}[h]
\caption{Neologism Fuzzy Metric}\label{algo:neo_fuzzy_metric}
\begin{algorithmic}[1]
\Require Source text $x$, generated translation $\hat{y}$, the set of corresponding spans of neologisms in $y^\subtxt{ref}$: $\mathcal{S}$, fuzzy string matching function $\mathcal{F}$.
\Ensure Neologism fuzzy score: \textsc{Fuzzy}.

\State Initialize count: $c \gets 0$.
\State Initialize the number of spans in $\mathcal{S}$: $s = \lvert \mathcal{S} \rvert$.
\For{$s \in \mathcal{S}$} 

    \If{$\mathcal{F}(s, \hat{y})$ is true}
        \State Increment count $c \gets c + 1$.
    \EndIf
\EndFor
\State \textbf{return} neologism fuzzy matching score: $c/s$.
\end{algorithmic}
\end{algorithm}

\begin{algorithm}[h]
\caption{Neologism Lem-Exact Metric}\label{algo:neo_lem_regex_metric}
\begin{algorithmic}[1]
\Require Source text $x$, generated translation $\hat{y}$, the set of corresponding spans of neologisms in $y^\subtxt{ref}$: $\mathcal{S}$, lemmatization function $\mathcal{L}$, exact string matching function $\mathcal{M}$.
\Ensure Neologism lem-exact score.
\State Initialize count: $c \gets 0$.
\State Initialize the number of spans in $\mathcal{S}$: $s = \lvert \mathcal{S} \rvert$.
\State Lemmatize $\hat{y}$ to obtain $\hat{y}_{l}$: $\hat{y}_{l} \gets \mathcal{L}(\hat{y})$.
\For{$s \in \mathcal{S}$} 
    \State Lemmatize $s$ to obtain $s_l$: $s_l \gets \mathcal{L}(s)$.
    \If{$\mathcal{M}(s_l, \hat{y}_{l})$ is true}
        \State Increment count $c \gets c + 1$.
    \EndIf
\EndFor
\State \textbf{return} neologism lem-exact score: $c/s$.
\end{algorithmic}
\end{algorithm}

\begin{algorithm}[h]
\caption{Neologism Lem-Fuzzy Metric}\label{algo:neo_lem_fuzzy_metric}
\begin{algorithmic}[1]
\Require Source text $x$, generated translation $\hat{y}$, the set of corresponding spans of neologisms in $y^\subtxt{ref}$: $\mathcal{S}$, lemmatization function $\mathcal{L}$, fuzzy matching function $\mathcal{F}$.
\Ensure Neologism lem-fuzzy score.
\State Initialize count: $c \gets 0$.
\State Initialize the number of spans in $\mathcal{S}$: $s = \lvert \mathcal{S} \rvert$.
\State Lemmatize $\hat{y}$ to obtain $\hat{y}_{l}$: $\hat{y}_{l} \gets \mathcal{L}(\hat{y})$.
\For{$s \in \mathcal{S}$} 
    \State Lemmatize $s$ to obtain $s_l$: $s_l \gets \mathcal{L}(s)$.
    \If{$\mathcal{F}(s_l, \hat{y}_{l})$ is true}
        \State Increment count $c \gets c + 1$.
    \EndIf
\EndFor
\State \textbf{return} neologism lem-fuzzy score: $c/s$.
\end{algorithmic}
\end{algorithm}

% \section{Baselines}
% \paragraph{General-purpose Models.}

% \paragraph{Machine-Translation Specific Models.}

% \paragraph{Supervised Fine-Tuning}

% \paragraph{RAG}

\section{Examples}
\label{app:ex}
In this section, we provide some detailed examples of outputs generated by \modelname{}.
Table~\ref{tab:find_new_meaning_example_our_model_gta} and~\ref{tab:find_new_meaning_example_our_model0} present \textbf{finding new meaning pattern} examples.
Table~\ref{tab:confirm_pattern_example_model} shows an example of the \textbf{finding new meaning pattern}.

\begin{table*}[h]
\begin{tcolorbox}[enhanced,
drop fuzzy shadow,]
\begin{CJK*}{UTF8}{gkai}
{\bf \modelname{} Instruction} 
You are a professional Chinese to English translator. Please translate the following text from Chinese to English. You must conduct reasoning inside <think> and </think> first.

[\dots]

<translation> final translation here </translation>. 

\textbf{Input} Chinese text: 尽管Rockstar是一家经验丰富而且成功的游戏公司，但三款标志性《给她爱》游戏的重制版因其艺术风格和性能而受到批评……
\end{CJK*}
\tcblower
\begin{CJK*}{UTF8}{gkai}
{\bf Output} <think> Okay, I need to translate the given Chinese sentence into English. Let's start by understanding the original text. 
The sentence is:  "尽管Rockstar是一家经验丰富而且成功的游戏公司，但三款标志性《给她爱》游戏的重制版因其艺术风格和性能而受到批评……"  
First, breaking it down.  
"尽管" translates to  "Although" or  "Despite".  
"Rockstar是一家经验丰富而且成功的游戏公司" –  "Rockstar is a seasoned and successful game company".  
"但" here is  "but".
Then,  "三款标志性《给她爱》游戏的重制版" –  "the three landmark 'Give Her Love' games' remasters " or "remake versions". 
"因其艺术风格和性能而受到批评" – "have been criticized for their artistic style and performance". 
The ellipsis at the end might indicate the sentence is continuing, but maybe it's just a pause.  
Wait, "《给她爱》" is the title of the games. 
Let me check if that's the correct translation. "给她爱" could be "Give Her Love" or maybe "Love Her" – but perhaps the actual game title is different. 
Maybe I should verify that. Let me search for "《给她爱》游戏" to confirm the English title.  

<search>《给她爱》游戏的英文名</search>  

<information>

\begin{CJK*}{UTF8}{bkai}
Doc 1(Title: Word: 給她愛) 給她愛 Part-of-speech: name Etymology: From English Grand Theft Auto → initialism GTA → pinyin Gěitā'ài. 
[\dots]
Word senses: 1. (Mainland China, neologism, euphemistic) Grand Theft Auto (video game franchise published by Rockstar Games) 
\end{CJK*}

% \begin{CJK*}{UTF8}{min}
Doc 2(Title: Word: otome game) otome game (plural otome games) Part-of-speech: noun Etymology: Borrowed from Japanese 乙({\CJKfamily{min}おと})女({\CJKfamily{min}め}){\CJKfamily{min}ゲーム} (otome gēmu, literally “girl game”). Word senses: 1. A story-based video game in which the player attempts to establish a romantic relationship between the female player character and one of several male characters.  **Non-disambiguated translation entries**: Sense: romance simulation game (usually the protagonist is a woman) - Japanese: 乙女{\CJKfamily{min}ゲーム} 
% \end{CJK*}

[\dots]

% Doc 3(Title: Word: 姫プレイ) 姫(ひめ)プレイ • (hime-purei) Part-of-speech: noun Etymology: From 姫(ひめ) (hime) + プレイ (purei). Word senses: 1. (online gaming, slang, of a female player) allowing male players to simp for her 
% Doc 4(Title: Word: play for love) play for love (third-person singular simple present plays for love, present participle playing for love, simple past and past participle played for love) Part-of-speech: verb Word senses: 1. (idiomatic) To play a game of cards without stakes.  **Non-disambiguated translation entries**: Sense: To play a game of cards without stakes - Polish (imperfective): grać dla zabawy - Russian: игра́ть на интере́с Doc 5(Title: Word: Zelda) Zelda or Zelda (plural Zeldas) Part-of-speech: name Etymology: Shortened from Griselda. Word senses: 1. (video games) Ellipsis of Legend of Zelda.  **Non-disambiguated translation entries**: Sense: female given name - French: Zelda - German (feminine): Zelda - Hebrew: זֶלְדָּה - Japanese: ゼルダ - Russian: Зельда - Russian: Зельде - Tamil: செல்டா - Ukrainian: Зелда  

</information>   

Hmm, so "《给她爱》" is mentioned in the context of being a game. The first document says that "给她爱" is a neologism in China for Grand Theft Auto (GTA), specifically GTA V, since China banned GTA V. 
But the original sentence mentions three landmark games. Wait, Rockstar Games has several titles like GTA III, GTA IV, GTA V, etc. Maybe the user is referring to three of their flagship games. 
But the original text says "《给她爱》游戏", which according to the search result, is a term used in China for GTA. 

[\dots]

But since the search shows that "给她爱" is a term for GTA, perhaps that's the case.   

% So, the translation would be: "Although Rockstar is a seasoned and successful game company, the remasters of three landmark GTA games have been criticized for their artistic style and performance..."   

[\dots]

I think that's the right approach. So the final translation would use "GTA" instead of "Give Her Love". </think>  

<translation>Although Rockstar is a seasoned and successful game company, the remasters of three landmark GTA games have been criticized for their artistic style and performance.</translation> \\
\end{CJK*}
\end{tcolorbox}

\caption{\label{tab:find_new_meaning_example_our_model_gta} Instruction, input and output of a \textbf{finding new meaning pattern} example when \modelname{} translates the Chinese text that contains a neologism. In this example, \modelname{} actively searches the neologism, finds the new meaning of the word, and reduces the ambiguous meaning of the word successfully by looking up a dictionary. }
\end{table*}

\begin{table*}[h]
\begin{tcolorbox}[enhanced,
drop fuzzy shadow,]
\begin{CJK*}{UTF8}{bkai}
{\bf \modelname{} Instruction} 
You are a professional Chinese to English translator. Please translate the following text from Chinese to English. You must conduct reasoning inside <think> and </think> first.

[\dots]

<translation> final translation here </translation>. 

\textbf{Input} Chinese text: 奧利給, 幹了兄弟們！
\end{CJK*}
\tcblower
\begin{CJK*}{UTF8}{bkai}
{\bf Output} <think>Okay, I need to translate the Chinese text "奧利給, 幹了兄弟們！" into English. Let's start by breaking down the sentence.
First, "奧利給" – that's probably a name. "奧利" could be ``Oli'' or "Aoli", and "給" here might be part of a phrase. 

Wait, maybe it's "奧利給" as a name, like "Ollie" or "Aolie". 

Then "幹了兄弟們！" – "幹了" could be "go" or "do" something. "幹了" might be an exclamation. "兄弟們" is "brothers" or "guys". 

Wait, "幹了" could be part of a slang or a phrase. 

Maybe "幹了" here is "Let's go" or "Do it". 

So maybe the whole phrase is "Ollie, let's go, brothers!" or "Ollie, do it, guys!" Hmm. Alternatively, "幹了" could be "干了" which is "do" or "carry out". So maybe "Ollie, let's do it, brothers!" or "Ollie, go for it, guys!" 
I should check if "奧利給" is a known name or if there's a common translation. Let me search for "奧利給" in Chinese to English.

<search>奧利給~英文名</search> 

<information>

[\dots]

% Doc 1(Title: Word: 奧利給) 奧利給 Part-of-speech: noun Etymology: Reversal of 給力噢 \begin{CJK*}{UTF8}{gbsn}/给力噢\end{CJK*} (gěilì ō, “come on!”). Popularized through a Kuaishou video uploaded by Huang Chunsheng (\begin{CJK*}{UTF8}{gbsn}黄春生\end{CJK*}), known as Chaoyang Winter Swimming Strange Pigeon on the Internet (with original sentence quoted below). The meaning of feces derives from video creator Daoshilaoba (\begin{CJK*}{UTF8}{gbsn}岛市老八\end{CJK*}), who mimicked the former quote before livestreaming coprophagy. Word senses: 1. (neologism, slang) feces 

Doc 2(Title: Word: 奧利給) 奧利給 Part-of-speech: intj Etymology: Reversal of 給力噢 \begin{CJK*}{UTF8}{gkai}/给力噢\end{CJK*} (gěilì ō, “come on!”). Popularized through a Kuaishou video uploaded by Huang Chunsheng (\begin{CJK*}{UTF8}{gkai}黄春生\end{CJK*}), known as Chaoyang Winter Swimming Strange Pigeon on the Internet (with original sentence quoted below). The meaning of feces derives from video creator Daoshilaoba (\begin{CJK*}{UTF8}{gkai}岛市老八\end{CJK*}), who mimicked the former quote before livestreaming coprophagy. Word senses: 1. (neologism, slang) come on!; go for it!; go! 

% Doc 3(Title: Word: Oliwier) Oliwier m Part-of-speech: name Etymology: Borrowed from French Olivier, from Latin Oliverius. Word senses: 1. a male given name, equivalent to English Oliver 

[\dots]

% Doc 4(Title: Word: Oliwer) Oliwer m Part-of-speech: name Etymology: Borrowed from French Olivier, from Latin Oliverius. Word senses: 1. a male given name, equivalent to English Oliver 
% Doc 5(Title: Word: 奧利維婭) 奧利維婭 Part-of-speech: name Word senses: 1. a transliteration of the English female given name Olivia  

</information>  

Okay, so  "奧利給 " seems to be a slang term derived from  "給力噢 " (gěilì ō), which means  "come on!" or  "go for it!". 
% The original phrase was reversed to  "奧利給" (Aoli Ji), which is used as an exclamation. So "奧利給" here is probably "Come on!" or "Go!".  Then "幹了兄弟們！" – "幹了" could be "gàn le", which literally means "done" or "finished", but in slang, "幹了" can be used as an exclamation similar to "Let's do it!" or "Let's go!" So "幹了兄弟們" would be "Let's go, brothers!" or "Let's do it, guys!".  Putting it all together: "奧利給，幹了兄弟們！" could be "Oli Ji, let's go, brothers!" but since "奧利給" is an exclamation, maybe "Come on, brothers! Let's do it!" or "Oli Ji, go for it, guys!" Wait, but the search results show that "奧利給" is a slang for "come on!" so maybe the translation is "Come on, brothers! Let's do it!" or "Come on, guys! Let's go!"   Alternatively, maybe "Oli Ji, let's go, brothers!" but since "奧利給" is a slang term, perhaps the name is "Oli Ji" but that might not be necessary. Maybe the phrase is "Oli Ji, come on, brothers!" or "Oli Ji, let's go, brothers!"   Wait, the original sentence is "奧利給，幹了兄弟們！" So it's two parts: "奧利給" followed by "幹了兄弟們". So maybe "Oli Ji, come on, brothers!" or "Oli Ji, let's go, brothers!"   But considering the search results, "奧利給" is an exclamation meaning "come on!" or "go for it!". So maybe the translation is "Come on, brothers! Let's do it!" or "Come on, guys! Let's go!"   
[\dots]
Alternatively, if "奧利給" is a name, then "Oli Ji, let's go, brothers!" but the search shows that "奧利給" is a slang term. So perhaps the translation is "Come on, brothers! Let's do it!" or "Come on, guys! Let's go!"   I think the best translation would be "Come on, brothers! Let's do it!" or "Come on, guys! Let's go!"   Wait, "幹了" could also be "gàn le", which is "done", but in this context, it's an exclamation. So maybe "Let's go, brothers!" or "Let's do it, guys!"   So combining "奧利給" (come on!) and "幹了兄弟們" (let's go, brothers!), the translation could be "Come on, brothers! Let's go!" or "Come on, guys! Let's do it!"   I think the most accurate translation based on the search results is "Come on, brothers! Let's go!" or "Come on, guys! Let's go!"   Alternatively, maybe "Oli Ji, come on, brothers!" but since "奧利給" is a slang term, perhaps the name is "Oli Ji" but that might not be necessary. The original sentence might be a call to action, so "Come on, brothers! Let's go!" or "Let's do it, guys!"   I'll go with "Come on, brothers! Let's go!" as the translation. </think> 

<translation>Come on, brothers! Let's go!</translation> \\
\end{CJK*}
\end{tcolorbox}

\caption{\label{tab:find_new_meaning_example_our_model0} Instruction, input and output of a \textbf{finding new meaning pattern} example when \modelname{} translates the Chinese text that contains a neologism. In this example, \modelname{} actively searches the neologism, finds the new meaning of the word, and reduces the ambiguous meaning of the word successfully by looking up a dictionary. }
\end{table*}

% \begin{table*}[h!]
% \begin{tcolorbox}[enhanced,
% drop fuzzy shadow,]
% \begin{CJK*}{UTF8}{bsmi}
% {\bf \modelname{} Instruction} 
% You are a professional Chinese to English translator. Please translate the following text from Chinese to English. You must conduct reasoning inside <think> and </think> first.

% \dots

% <translation> final translation here </translation>. 

% \textbf{Input} Chinese text: lo娘
% \end{CJK*}
% \tcblower
% \begin{CJK*}{UTF8}{bsmi}
% {\bf Output} <think>Okay \\
% \end{CJK*}
% \end{tcolorbox}

% \caption{\label{tab:example_our_model} \textbf{Finding new meaning pattern}: Instruction, input and output of our agentic model when translating the Chinese text that contains a neologism. In this example, \modelname{} actively searches the neologism. finds the new meaning of the word, and reduces the ambiguous meaning of the word successfully by looking up dictionary items. }
% \end{table*}

\begin{table*}[h]
\begin{tcolorbox}[enhanced,
drop fuzzy shadow,]
\begin{CJK*}{UTF8}{bkai}
{\bf \modelname{} Instruction} 
You are a professional Chinese to English translator. Please translate the following text from Chinese to English. You must conduct reasoning inside <think> and </think> first.

[\dots]

<translation> final translation here </translation>. 

\textbf{Input} Chinese text: 視頻來源：優兔
\end{CJK*}
\tcblower
\begin{CJK*}{UTF8}{bkai}
{\bf Output} <think> Okay, I need to translate "視頻來源：優兔" from Chinese to English. 
Let's start by breaking down the sentence. "視頻" is "video", "來源" is "source" or "from". 
"優兔" is probably a proper noun, maybe a platform or a name. Wait, "優兔" could be "YouTu" or maybe it's a misspelling of "YouTube". 
Oh, right, YouTube's Chinese name is sometimes written as "优兔" or "YouTube".
So "視頻來源：優兔" would be "Video source: YouTu" or "Video source: YouTube".
But I should confirm if "優兔" is indeed YouTube. 
Let me check that.

<search>什么是優兔</search>

<information> 

Doc 1(Title: Word: 優兔) 優兔 Part-of-speech: name Etymology: Borrowed from English YouTube. Word senses: 1. (neologism) YouTube 

Doc 2(Title: Word: 玉兔) 玉兔 Part-of-speech: noun Word senses: 1. (Chinese mythology) The Jade Rabbit; alternative name for 月兔 (yuètù, “the Moon Rabbit”).  2. (literary, figuratively) the moon 

Doc 3(Title: Word: 優輝) 優(\begin{CJK}{UTF8}{min}ゆう\end{CJK})輝(\begin{CJK}{UTF8}{min}き\end{CJK}) • (Yūki) Part-of-speech: name Word senses: 1. a male given name 

Doc 4(Title: Word: 優也) 優(\begin{CJK}{UTF8}{min}ゆう\end{CJK})也(\begin{CJK}{UTF8}{min}や\end{CJK}) • (Yūya) Part-of-speech: name Word senses: 1. a male given name 

% Doc 5(Title: Word: 大白兔) 大白兔 Part-of-speech: noun Etymology: Sense \#2 is influenced by “White Rabbit”, a brand of milk candy in China. Word senses: 1. (literally) big white rabbit  2. (figurative, slang) big breast (that bounces/jumps up and down as the person moves around)  
[\dots]

</information>

Okay, so "優兔" is mentioned as a neologism for YouTube. So the translation should be "Video source: YouTube". But maybe the original text is using "優兔" as the name, so perhaps it's better to keep it as "YouTu" or "You Tube". Wait, the search result says "優兔" is a borrowed term from YouTube. So the correct translation would be "Video source: YouTube". Alternatively, if they want to keep the original name, maybe "Video source: YouTu". But since the search says it's YouTube, I should go with that. So the final translation would be "Video source: YouTube". Let me check again. The original text is "視頻來源：優兔", so "视频来源" is "video source" or "source of the video". "优兔" is YouTube. So the translation is "Video source: YouTube" or "Video source: YouTu". But according to the search, "优兔" is YouTube. So probably "Video source: YouTube". Yeah, that makes sense. </think>  

<translation>Video source: YouTube</translation> \\
\end{CJK*}
\end{tcolorbox}

\caption{\label{tab:confirm_pattern_example_model}  Instruction, input and output of a \textbf{confirmation pattern} example when \modelname{} translates the Chinese text that contains a neologism. In this example, \modelname{} makes a correct assumption about the neologism, and use the search tool to find and verify the meaning of the word by looking up dictionary items.}
\end{table*}

\section{Construction Details of \datasetname{} Dataset}
\subsection{Construction Process.}
\label{app:neko_cons_details}
We use the raw Wiktionary data version (19.6GB)~\citep{ylonen-2022-wiktextract} dated 2025-08-23\footnote{\url{https://kaikki.org/dictionary/rawdata.html}} for extracting word entries.

Table~\ref{tab:neko_train_translation_prompt} presents the prompt that we use to translate the example sentences in the \textbf{\textit{Type 2}} word entries of \datasetname{} dataset. We sample 700 examples from the \textbf{\textit{Type 2}} word entries and translate the source text into the other 15 languages, totaling 10,425 translation pairs for training. Note that we provide the definitions of neologisms to make the translation process easier for \gptfivename{}. 

\begin{table}[h]
\centering
\resizebox{0.45\textwidth}{!}{%
    \begin{tabular}{p{8cm}}
        \toprule
        % Inference Prompt                                        \\ \midrule
You are a professional \{src\_lang\} to \{tgt\_lang\} translator. \\
I will provide you with the part-of-speech and gloss information of some words. Please use this information to translate the sentence. \\
Word: \{word\} \\
Part-of-speech: \{pos\} \\
Glosses: \{glosses\} \\
Please conduct the reasoning process first, and subsequently present the finalized translation enclosed within the tags: <translation> final translation </translation>. \\
\{src\_lang\} text: \{src\_text\}. \\
        \bottomrule                                                     
    \end{tabular}%
}
\caption{Translation prompt used to obtain translations of example sentences in the train set of \datasetname{} dataset. The model that we use for translation is \gptfivename{}.}
\label{tab:neko_train_translation_prompt}
\end{table}

Table~\ref{tab:neko_alignment_prompt} shows the prompt that use for finding the corresponding spans in the target sentences for the neologisms in the source sentences. Note that we provided additional information about the neologisms, such as definitions and the part-of-speech tags to make the alignment process easier.
For one neologism that might correspond to multiple words in the other language, we annotate the word that most reflects the meaning of the neologism.
The reason why we use \gptfivename{} rather than word alignment models is that these neologisms are relatively new and may not be seen in the training process of the current word alignment models and we can provide additional information about the neologisms in the prompt of \gptfivename{} to make the alignment process more accurate, which we cannot do if we use word alignment models.
After obtaining the corresponding spans of neologisms, we employ human volunteers to validate and refine the generated spans for Japanese-English and Chinese-English pairs. We find that the quality of spans generated by \gptfivename{} is quite high and there is almost no need to modify them.

\begin{table}[h]
\centering
\resizebox{0.45\textwidth}{!}{%
    \begin{tabular}{p{8cm}}
        \toprule
        % Inference Prompt                                        \\ \midrule
You are a professional word aligner. \\
I will provide a translation pair along with the part-of-speech and gloss information for a specific word in \{src\_lang\}. \\
Your task is to identify the corresponding word in \{tgt\_lang\}. \\
\{src\_lang\} text: \{src\_text\} \\
\{tgt\_lang\} text: \{tgt\_text\} \\
\{src\_lang\} Word: \{word\} \\
Part-of-speech: \{pos\} \\
Glosses: \{glosses\} \\
Please identify the corresponding word in the \{tgt\_lang\} text. \\
Present the identified word enclosed within the tags: <aligned\_word> corresponding word </aligned\_word> \\
        \bottomrule                                                     
    \end{tabular}%
}
\caption{The prompt that we use for finding the corresponding spans in the target sentences for the neologisms in the source sentences. The model that we use for finding alignments is \gptfivename{}.}
\label{tab:neko_alignment_prompt}
\end{table}

\subsection{Human Validation on the Synthetic Translations of the Neologism for Type 2 data}
\label{app:validation_type_2}
The way of generating the synthetic translations is divided into two steps:

\begin{itemize}
    \item The first step was that we provided the details of the neologisms in the source sentences for gpt-5-2025-08-07 to generate translations. Therefore, it is relatively easy for gpt-5-2025-08-07 to translate the source sentences containing neologisms.
    \item The second step was using gpt-5-2025-08-07 to find the corresponding spans in the generated translations in step 1 given the source sentences, neologisms with part-of-speech and gloss information and generated translations.
\end{itemize}
We found that the synthetic translations using the above approach were of high quality, fluent and natural.

Specifically, we randomly sampled 200 English-Japanese and English-Chinese synthetic translations from our training data and conducted a human analysis to verify if the translations are accurate and convey the meaning of source sentences and neologisms or not.
We use the pass rate to check the quality with the result shown in Table~\ref{tab:pass_rate_train}.
The failure reasons and the error spans in the synthetic translations that did not pass human validation are shown in Table~\ref{tab:failture_case_in_train_set}.

\begin{table}[h]
    \centering
    \resizebox{0.25\textwidth}{!}{%
    \begin{tabular}{lc}
    \toprule
        Language-pair & Pass rate \\
    \midrule
        English-Japanese & 94.5\% \\
        English-Chinese & 97.5\% \\
    \bottomrule
    \end{tabular}}
    \caption{Pass rate of the training data by human validation.}
    \label{tab:pass_rate_train}
\end{table}

\begin{table*}[h]
    \centering
    \resizebox{0.8\textwidth}{!}{%
    \begin{tabular}{p{8cm}p{5cm}p{5cm}}
    \toprule
        Failure reason & English-Japanese & English-Chinese \\
    \midrule
        Copy paste without explanation (affecting the meaning) & Hu, HODL, hu, Chican@s & Melfie, MGTOW \\
    \midrule
        Surface meaning translation without explanation & {\CJKfamily{min}アマトノーマティブな, シューフィー, スウィフトボート式中傷, パルーザ, シット, レシオ}  & 泡泡糖坏女孩核, 解融合 \\
    \midrule
        Unusual Human Name Translation & - & Daniel -> 但以理 (More common translation is 丹尼尔) \\
    \midrule
        Language Mixing (Code-switching) & {\CJKfamily{min}websideマナー} & - \\
    \bottomrule
    \end{tabular}}
    \caption{Failure reasons and errors spans in the synthetic training set.}
    \label{tab:failture_case_in_train_set}
\end{table*}

\section{Statistics of \datasetname{} Dataset}
\label{app:statistics}
The statistics about the train, val and test splits of \datasetname{} dataset can be found in Table~\ref{tab:data-comparison},~\ref{tab:neko_statistics_val_test} and~\ref{tab:neko_statistics_train}.

\begin{table}[H]
\centering
\resizebox{0.3\textwidth}{!}{%
    \begin{tabular}{lll}
    \toprule
    Language pair & \# (val) & \# (test) \\
    \midrule
    zh-en & 61 & 578 \\
    ru-en & 4 & 38 \\
    fr-en & 1 & 20 \\
    uk-en & 2 & 17 \\
    ja-en & 7 & 20 \\
    de-en & 3 & 30 \\
    cs-en & - & 6 \\
    pl-en & 5 & 34 \\
    en-xx (ref-free) & - & 270 \\
    \bottomrule
    \end{tabular}
}
\caption{Number of examples grouped by language pairs in the \textbf{val} and \textbf{test} splits of the \datasetname{} dataset.}
\label{tab:neko_statistics_val_test}
\end{table}

\begin{table*}[h]
\centering
\resizebox{0.8\textwidth}{!}{%
    \begin{tabular}{p{0.8\textwidth}}
    \toprule
    Language pair: Number \\
    \midrule
    en-is,
    en-ru,
    en-km,
    en-ha,
    en-de,
    en-ja,
    en-cs,
    en-uk,
    en-ta,
    en-pl,
    en-fr,
    en-he,
    en-zh,
    en-hr,
    en-ps: 603 \\
    fr-en,
    fr-is,
    fr-ru,
    fr-km,
    fr-ha,
    fr-de,
    fr-ja,
    fr-cs,
    fr-uk,
    fr-ta,
    fr-pl,
    fr-he,
    fr-zh,
    fr-hr,
    fr-ps: 18 \\
    zh-en,
    zh-is,
    zh-ru,
    zh-km,
    zh-ha,
    zh-de,
    zh-ja,
    zh-cs,
    zh-uk,
    zh-ta,
    zh-pl,
    zh-fr,
    zh-he,
    zh-hr,
    zh-ps: 68 \\
    de-en,
    de-is,
    de-ru,
    de-km,
    de-ha,
    de-ja,
    de-cs,
    de-uk,
    de-ta,
    de-pl,
    de-fr,
    de-he,
    de-zh,
    de-hr,
    de-ps,
    is-en,
    is-ru,
    is-km,
    is-ha,
    is-de,
    is-ja,
    is-cs,
    is-uk,
    is-ta,
    is-pl,
    is-fr,
    is-he,
    is-zh,
    is-hr,
    is-ps: 3 \\
    \bottomrule
    \end{tabular}
}
\caption{The number of examples grouped by language pairs in the \textbf{train} split of the \datasetname{} dataset.}
\label{tab:neko_statistics_train}
\end{table*}

\section{More Analysis}
\label{app:more_analysis}
The following analysis is conducted using \modelname{}-8B.
\subsection{WMT24++}
\label{wmt24pp}
We also analyze the performance of our model on the general MT task using the subset of WMT24++ covered by our research language range. We remove the bad source example sentences and translations. Post-edited translations are utilized as reference translations.

\begin{figure}
    \centering
    \includegraphics[width=0.6\linewidth]{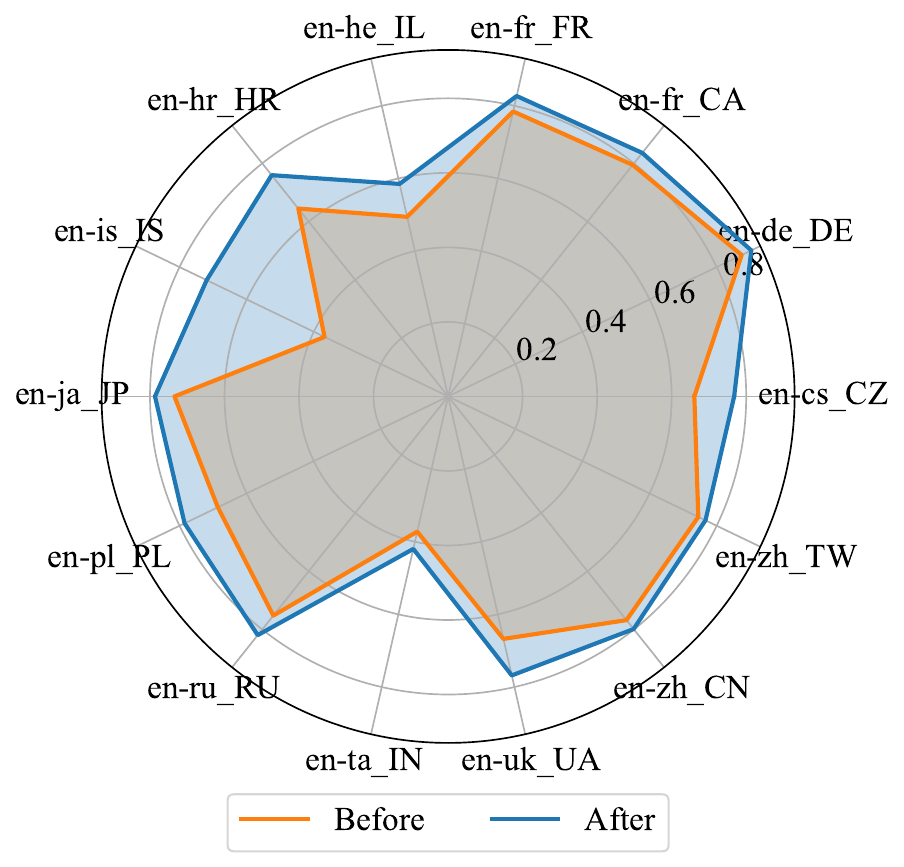}
    \caption{Comparison of XCOMET-XL scores on the WMT24pp test dataset grouped by language pairs between before and after our training using the training split of the \datasetname{} dataset.}
    \label{fig:wmt24pp_radar_plot_per_lang_pair}
\end{figure}

% WMT24++~\citep{deutsch-etal-2025-wmt24} dataset is utilized to assess the performance of \modelname{} on the general MT task. 
This evaluation is a type of out-of-domain setting because our model is trained on the \datasetname{} dataset.
% Table~\ref{tab:wmt24pp-result} shows the WMT24++ result on a subset that contains 14 language pairs covered in our research language range. 
Figure~\ref{fig:wmt24pp_radar_plot_per_lang_pair} presents the comparison results grouped by language pairs before and after the training using the \datasetname{} dataset. 

\subsection{Training Dynamics}
\begin{figure}[h]
    \centering
    \includegraphics[width=0.75\linewidth]{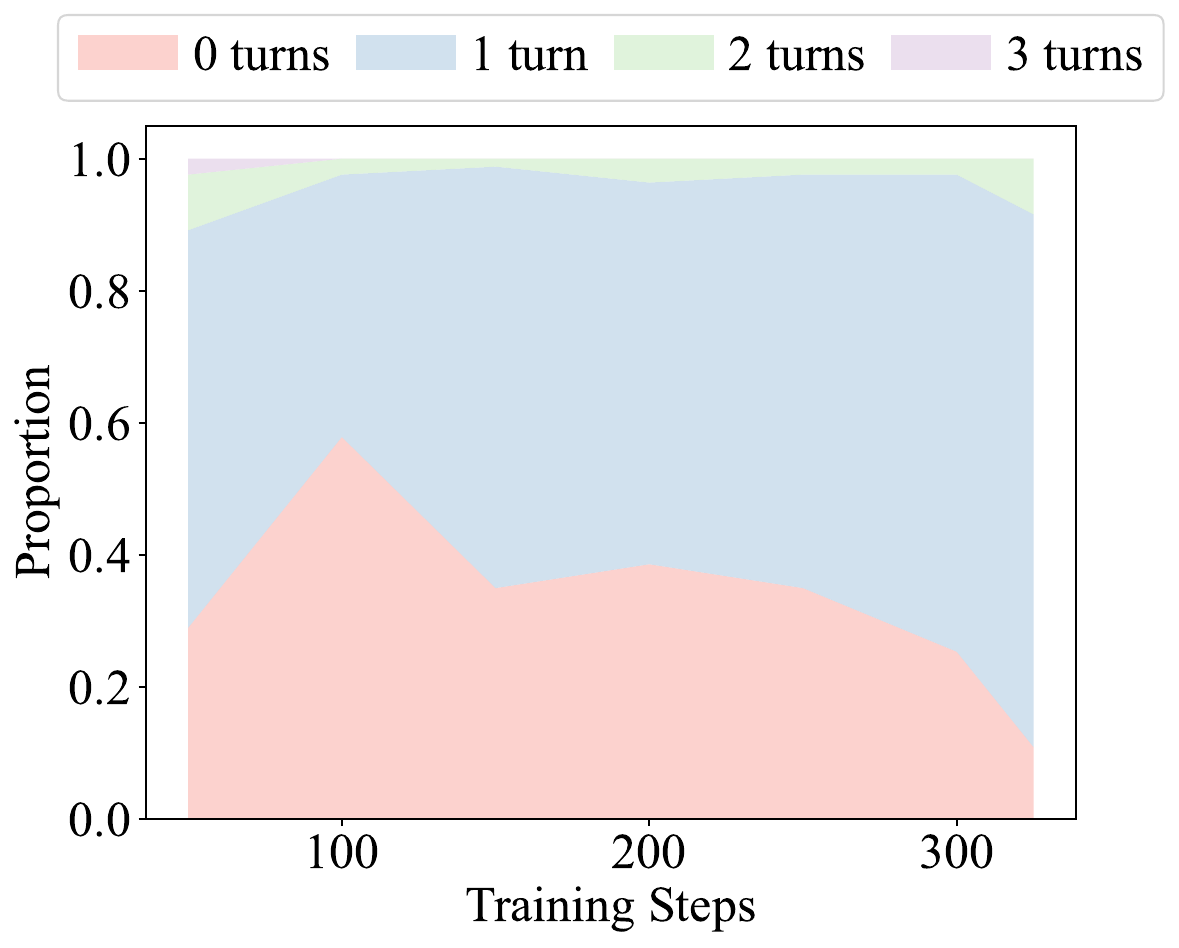}
    \caption{Change of the proportions about the search turns on the validation set over the course of training.}
    \label{fig:search_turn}
\end{figure}

We analyze the training dynamics of the number of search turns, over training steps.
The result of search turns of the validation set is presented in Figure~\ref{fig:search_turn}.
% The result of output length is presented in Figure~\ref{fig:avg_output}.
We find that (1) at the beginning, 30\% of the responses do not use searching and over the training steps, the proportion of not using searching first get larger and declines; (2) most of responses use searching once; (3) The proportion of not using search shows an opposite trend to the output length. This aligns with our intuition that using search tends to increase the output length.

\subsection{Case Studies}
\label{app:case_study}
Table~\ref{tab:case-study} shows some translation examples of \modelname{}-8B.
\begin{table*}[h]
\centering
\resizebox{0.7\textwidth}{!}{%
    \begin{tabular}{m{4cm}m{2cm}p{9cm}}
    \toprule
    Source text      &   Model  & Translation \\ 
    \midrule
    
    \multirow{3}{4cm}[-3.5em]{ 「我一直都在长草这部美剧啊，可惜太忙了没时间看。」 } 
    & {Qwen3-8B}                      &  "I've always wanted to watch the TV series 'The Office,' but I've been too busy to watch it." \\ \cline{2-3}
    & {Hunyuan-MT-7B}                 &  “I’ve been wanting to watch that amazing TV series ‘Grass’ for a long time, but unfortunately I’m too busy to find the time to watch it.” \\ \cline{2-3}
    & {\modelname{}}                  &   I've been really looking forward to this American TV show all along, but unfortunately, I've been too busy to watch it.           \\ \cline{1-3}

    \multirow{3}{4cm}[-1em]{ lo娘 } 
    & {Qwen3-8B}                      &  Lo娘 \\ \cline{2-3}
    & {Hunyuan-MT-7B}                 & lo娘 \\ \cline{2-3}
    & {\modelname{}}                  &   Lolita girl           \\ \cline{1-3}

    \multirow{3}{4cm}[-1.5em]{ 自己都会笑称自己系一个「铁胶」 } 
    & {Qwen3-8B}                      &  One would even laughingly call oneself a "steel glue" \\ \cline{2-3}
    & {Hunyuan-MT-7B}                 &  I even joke with myself about being a “glue stick” (a person who gets stuck to things easily). \\ \cline{2-3}
    & {\modelname{}}                  &   One would jokingly call oneself a 'railfan'.          \\ \cline{1-3}

    \multirow{3}{4cm}[-1em]{ 放黑枪 } 
    & {Qwen3-8B}                      &  Fire a shot in the dark \\ \cline{2-3}
    & {Hunyuan-MT-7B}                 &  Draw the black gun. \\ \cline{2-3}
    & {\modelname{}}                  &   to launch a covert attack          \\ 
    
    \bottomrule
    \end{tabular}%
}
\caption{Case studies.}
\label{tab:case-study}
\end{table*}

\subsection{XCOMET and MetricX24 Scores}
\label{app:xcomet-metricx}
XCOMET-XL and MetricX-24 scores of \modelname{}-8B are presented in Table~\ref{tab:xcomet-metricx24-result}.

\begin{table}[H]
\centering
% \small
\setlength{\tabcolsep}{4pt}
\renewcommand{\arraystretch}{1}
\resizebox{0.4\textwidth}{!}{%
    \begin{tabular}{l*{2}{c}}
        \toprule
        % \multirow{2.5}{*}{\sc Model} & \multicolumn{4}{c}{\sc Neologism-Specific} & & \multicolumn{3}{c}{\sc Overall Quality} \\
        % \cmidrule(lr){2-5} \cmidrule(lr){6-9}
        \textsc{Model} & \textsc{XCOMET} & \textsc{MetricX-24}  \\ 
        \midrule
        % \multicolumn{9}{c}{\textit{\textbf{Closed}}} \\
        % \multicolumn{9}{@{}l}{\textcolor{lightgray}{\textit{General Purpose LLMs}}} \\
        % {GPT-5}               & 27.19 & 29.48 & 30.96 & 31.49 & & & 77.38 & 85.58          \\
        % {Claude-Sonnet-4.5}   & 26.11 & 27.73 & 29.21 & 30.55 & & & 73.00 & 77.50              \\
        % {Gemini-2.5-Pro}      &  28.26 & 30.28 & 33.51 & 33.92 & & & 76.05 & 83.29         \\
        % {Qwen-Max} \\
        % {Grok4} \\
        % \multicolumn{10}{@{}l}{\textcolor{lightgray}{\textit{Translation-Specific LLMs}}} \\
        % {Qwen-MT-Plus} \\
        % {DeepL Translate} & 11.71 & 14.13 & 14.27 & 16.29 & \\
        % {Google Translate} & 21.80 & 23.28 & 25.57 & 25.30 &  \\
        % {GPT5-Nano} & 22.21 & 24.76 & 24.76 & 26.92\\
        % \midrule
        % \cdashlinelr{1-9}
        % \midrule
        % \multicolumn{9}{c}{\textit{\textbf{Open}}} \\
        \multicolumn{3}{@{}l}{\textcolor{lightgray}{\textit{General Purpose LLMs}}} \\
        % {DeepSeek-V3.2-Exp}             &       &       &      &            \\
        % {Kimi-K2-Instruct} & & & & \\
        % {GLM-4.6 (357B)} & & & & \\
        
        % {GPT-OSS-120B} \\
        % {Qwen3-235B-A22B-Instruct-2507} \\
        {Llama-3.1-8B-Instruct}  & 78.81 & 6.6  \\
        {Aya-23-8B}              & 79.27 & 6.3      \\
        {Aya-expanse-8B}         & 81.09 & 5.8      \\
        % {Gemma-3-12b-it} \\
        % {Gemma-3-27b-it} \\
        {Qwen3-8B}  & 81.12 & 5.7 \\
        % \multicolumn{10}{@{}l}{\textcolor{lightgray}{\textit{General Purpose LLMs (Above 10B)}}} \\
        {Qwen3-14B} & 83.36 & 5.6   \\
        {Qwen3-32B} & 83.33 & 5.2   \\
        {GPT-OSS-20B}  & 81.39 & 5.8   \\
        % {Llama-3.3-70B-Instruct} & 16.15 & 18.30 & 20.46 & 21.40 & & 84.39 & 5.1 \\
        % {DeepSeek-R1-Distill-Qwen-7B} \\
        % {Qwen3-Next-80B-A3B-Instruct}   & & & & \\
        % {Qwen3-8B (CoT)}           &       &       &      &              \\
        \multicolumn{3}{@{}l}{\textcolor{lightgray}{\textit{Translation-Specific LLMs}}} \\
        % {ALMA} \\
        {ALMA-R-7B} & 79.11 & 6.1 \\
        {MT-R1-Zero-7B-Mix}   & 79.86 & 6.0     \\
        {X-ALMA} &  82.88 & 5.4 \\
        {TowerInstruct-7B-v0.2}  & 78.03 & 6.6   \\
        % {Tower-Plus-9B}   \\
        
        % {SSR-Zero} \\
        % {TAT-R1} \\
        % {DeepTrans} \\
        % {DRT} \\
        % {ExTrans-7B} \\
        {mExTrans-7B} & 80.81 & 5.7 \\
        % {GemmaX2-28-9B-v0.1} & 83.35 & 5.5 \\
        {Hunyuan-MT-7B}  & 82.36 & 5.2  \\
        \cdashlinelr{1-3}
        
        % \multicolumn{8}{c}{\textit{\textbf{Ours}}} \\
        \multicolumn{3}{@{}l}{\textcolor{lightgray}{\textit{Qwen3-8B}}} \\
        {SFT}       & 80.60 & 5.6  \\
        {RAG}       &  75.42 & 6.7 \\
        % {RAG (definitions$+$example sentences)}  & 23.96      & 24.76      &26.92      & 27.86              \\
        % {RAG (oracle)}           &   &       &      &               \\
        % \multicolumn{10}{@{}l}{\textcolor{lightgray}{\textit{w/o. Search Reward and training with definitions}}} \\

        % GRPO wo search
        {GRPO}      & 82.30 & 5.5  \\ 
        % {Ours (GRPO, definitions only)} & 20.19 & 23.15 & 23.01 & 25.17 & & 82.63 & 52.76 & 61.69   \\
        {\modelname{}} & \textbf{83.42} & \textbf{5.1}    \\ 
        % {Ours (GRPO, definitions$+$example sentences)} & 21.27 & 23.28 & 24.90 & 26.38 & & 82.51  \\
        % {Ours (RQE-GRPO, definitions$+$example sentences)}   & 23.82 & 26.24 & 27.05 & 28.94 & & 83.28 &   &     \\ 
        % {Ours (GRPO, oracle)}  & 36.47 & 38.49 & 41.18 & 41.72 & & 83.83 &        \\ 
        % {Ours (RQE-GRPO, oracle)}    & \textbf{44.41} & \textbf{46.57} & \textbf{48.72} & \textbf{49.80} & & \textbf{84.41} &        \\ 
        % \multicolumn{10}{@{}l}{\textcolor{lightgray}{\textit{w. Search Reward}}} \\
        % {Ours (w. rqe, definitions only)}   &  & &  &  & &  & &     \\ 
        \bottomrule
    \end{tabular}%
}
\caption{Main results of XCOMET and MetricX-24.}
\label{tab:xcomet-metricx24-result}
\end{table}

\subsection{LLM-as-a-judge Evaluation by Language Pair}
\label{app:llm_judge_lang_pair}
\begin{figure}[h]
     \centering
     \begin{subfigure}[b]{0.21\textwidth}
         \centering
         \includegraphics[width=\textwidth]{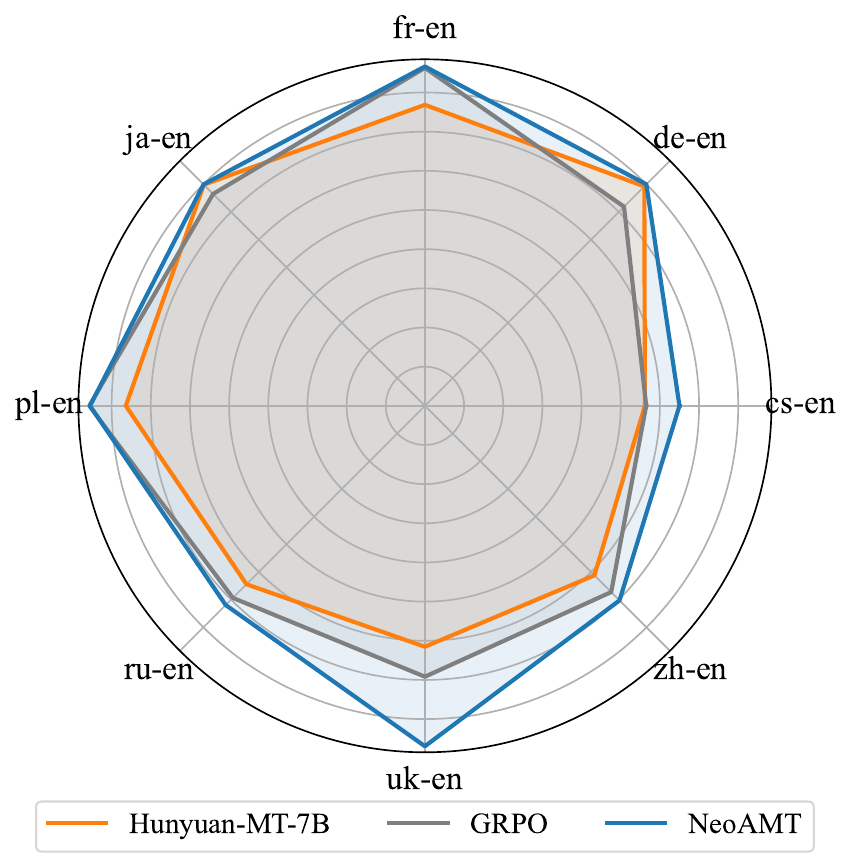}
         \caption{GEMBA score per lang pair.}
         \label{fig:radar_plot_per_lang_gemba}
     \end{subfigure}
     \hfill
     \begin{subfigure}[b]{0.21\textwidth}
         \centering
         \includegraphics[width=\textwidth]{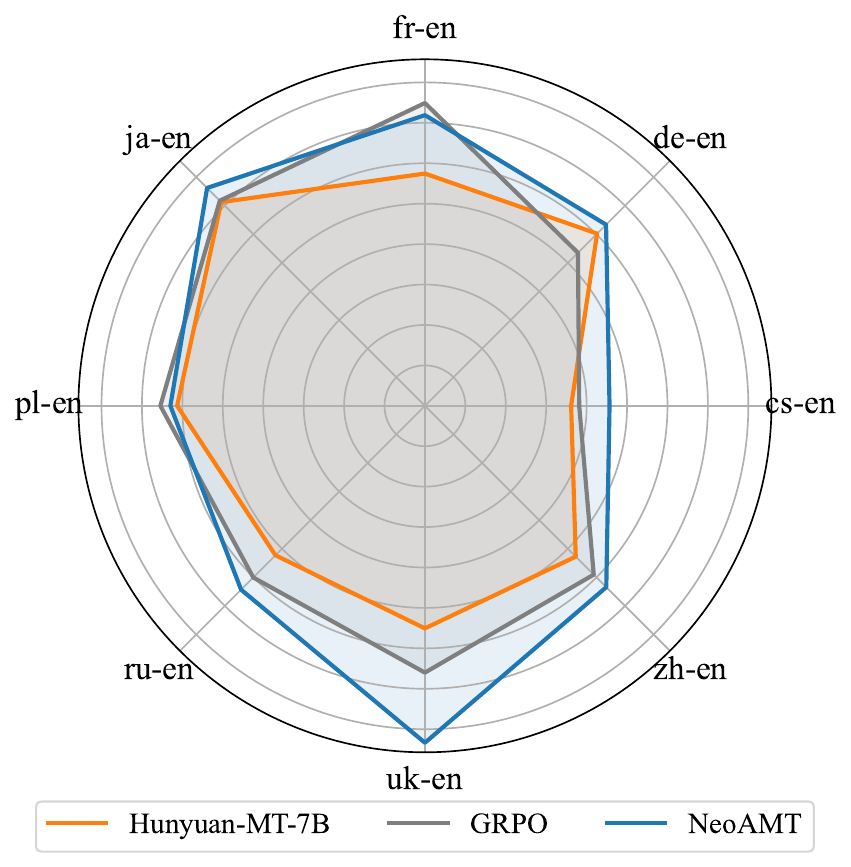}
         \caption{LJ(GPT5) score per lang pair.}
         \label{fig:radar_plot_per_lang_lj_gpt5}
     \end{subfigure}
    \caption{LLM-as-a-judge results grouped by language pairs.}
    \label{fig:radar_plot_per_lang}
\end{figure}

Table~\ref{fig:radar_plot_per_lang} shows the LLM-as-a-judge scores of \modelname{}-8B on the test split of the \datasetname{} dataset.

\subsection{Ablation study of RAG}
\label{app:rag_neologism_specific}
Neologism-specific scores of the RAG ablation study are shown in Table~\ref{tab:rag_neologism_specific}.
\begin{table}[H]
    \centering
    \resizebox{0.48\textwidth}{!}{%
    \begin{tabular}{l*{4}{c}}
    \toprule
    \textsc{MODEL} & \textsc{EXACT} & \textsc{FUZZY} & \textsc{LEM-EXACT} &  \textsc{LEM-FUZZY}  \\
    \midrule
        {Qwen3-4B + RAG}  &  30.95 & 27.72 & 34.58 & 31.22 \\
        {\modelname{}-4B} & {17.63} & {19.11} & {20.05} & {21.53}  \\
        {Qwen3-8B + RAG}  & 23.68 & 23.14 & 26.64 & 25.43  \\
        {\modelname{}-8B} & 22.34 & {26.24} & {25.84} & {28.67}     \\  
    \bottomrule
    \end{tabular}}
    \caption{Neologism-specific scores about the RAG ablation study.}
    \label{tab:rag_neologism_specific}
\end{table}

\section{Screenshot of Translation Ranking Annotation Application.}
\label{app:screenshot}
\begin{figure}[h]
    \centering
    \includegraphics[width=0.45\textwidth]{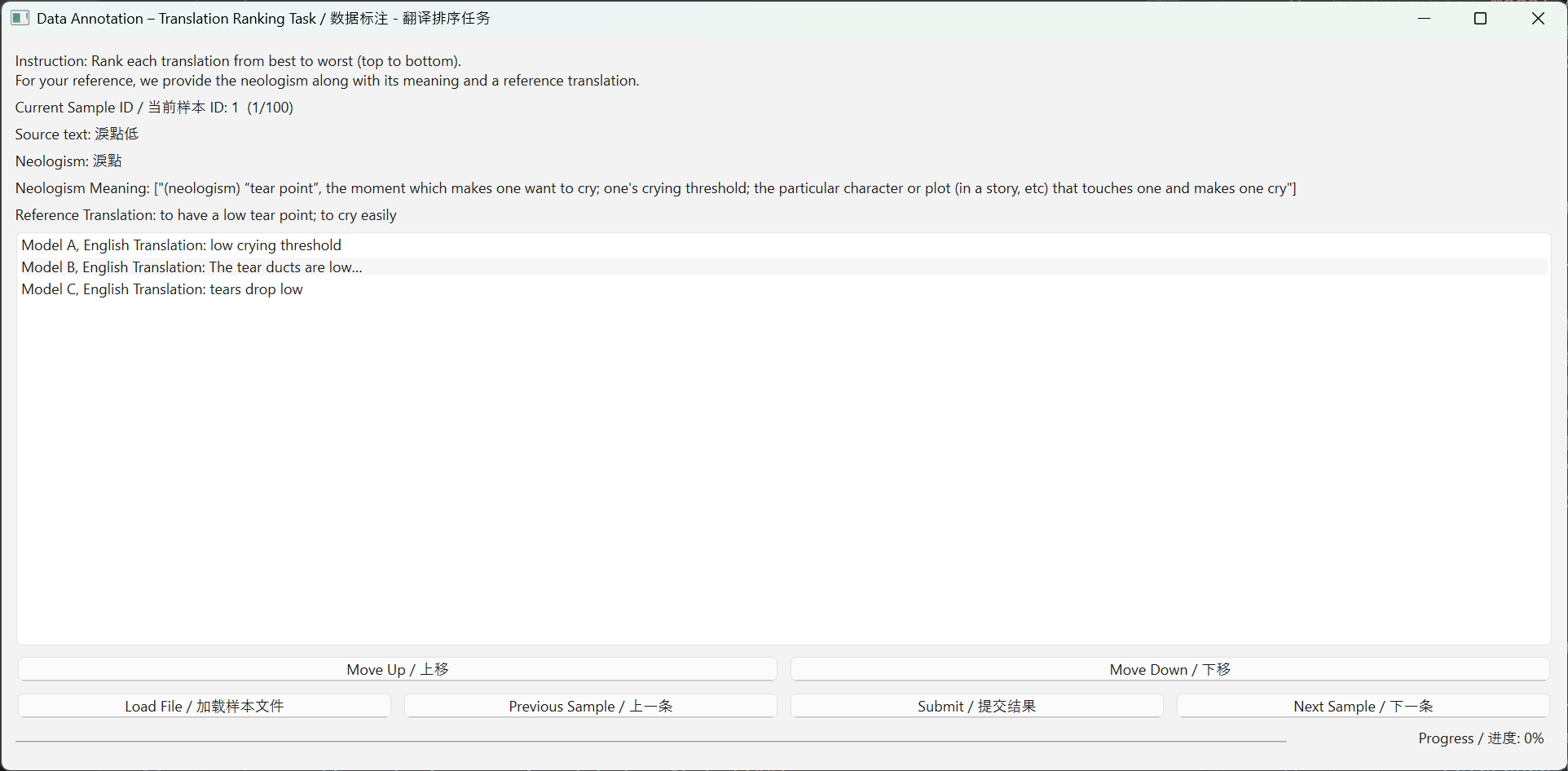}
    \caption{Screenshot of translation ranking annotation application.}
    \label{fig:ranking_app_screenshot}
\end{figure}

Human volunteers use an in-house application to rank translations. The screenshot of the application is shown in Figure~\ref{fig:ranking_app_screenshot}.

\end{CJK*}
\end{document}